\documentclass[11pt]{article}

\PassOptionsToPackage{table}{xcolor}
\usepackage[preprint]{acl}

\usepackage{times}
\usepackage{latexsym}
\usepackage[T1]{fontenc}
\usepackage[utf8]{inputenc}
\usepackage{microtype}
\usepackage{inconsolata}
\usepackage{graphicx}
\usepackage{amsmath}
\usepackage{amssymb}
\usepackage{booktabs}
\usepackage{multirow}
\usepackage{float}
\usepackage{tikz}

\definecolor{ourblue}{rgb}{0.368,0.507,0.71}
\definecolor{groupyellow}{RGB}{255,247,214}
\definecolor{extsand}{HTML}{EFD092}
\definecolor{tgtlav}{HTML}{C7BFF5}
\hypersetup{citecolor=ourblue, linkcolor=ourblue, urlcolor=ourblue}

\newcommand{\MAT}{\mathrm{MAT}}


\title{Target Regeneration Shrinks the Gains of Draft-Training Objectives\\in Speculative Decoding}

\author{
  Jungseob Lee$^{1}$ \quad Hogun Seong$^{2}$ \quad
  Euijun Choi$^{2}$ \quad Sugyeong Eo$^{2\dagger}$ \\
  $^{1}$Korea University \quad $^{2}$Yonsei University Mirae Campus \\
  {\small\texttt{omanma1928@korea.ac.kr}} \\
  {\small\texttt{\{tjf8200,euijun1015,s.eo\}@yonsei.ac.kr}} \\
  {\small $\dagger$Corresponding author.}
}

\begin{document}
\maketitle

\begin{abstract}
Speculative decoding accelerates large language models by verifying drafted tokens in parallel. New draft-training objectives are usually credited with the acceptance they add over cross-entropy on externally written responses, yet many pipelines let the target rewrite them. We ask how much of that gain survives. With a vision-language target and a block drafter, we train three objectives on four versions of one corpus: cross-entropy on corpus tokens, token-level distillation from the target, and distillation gated to positions that would survive verification. Distillation and the gate add large gains on external responses, but after greedy regeneration, distillation adds a smaller gain and the gate little or none. The shrinkage persists across data orders, position weights, verifiers, decoding temperatures and, for distillation, initializations and learning rates. Most of it remains at matched supervised length and at each objective's best measured training length. A second vision corpus shows the shrinkage, as does a text-only replication, for distillation also with an autoregressive drafter. Objective gains should therefore be measured against a target-regenerated baseline and across training lengths. Our code is available at \url{https://github.com/js-lee-AI/TargetRegeneration}.
\end{abstract}

\section{Introduction}

\begin{figure}[t]
\centering
\includegraphics[width=\columnwidth]{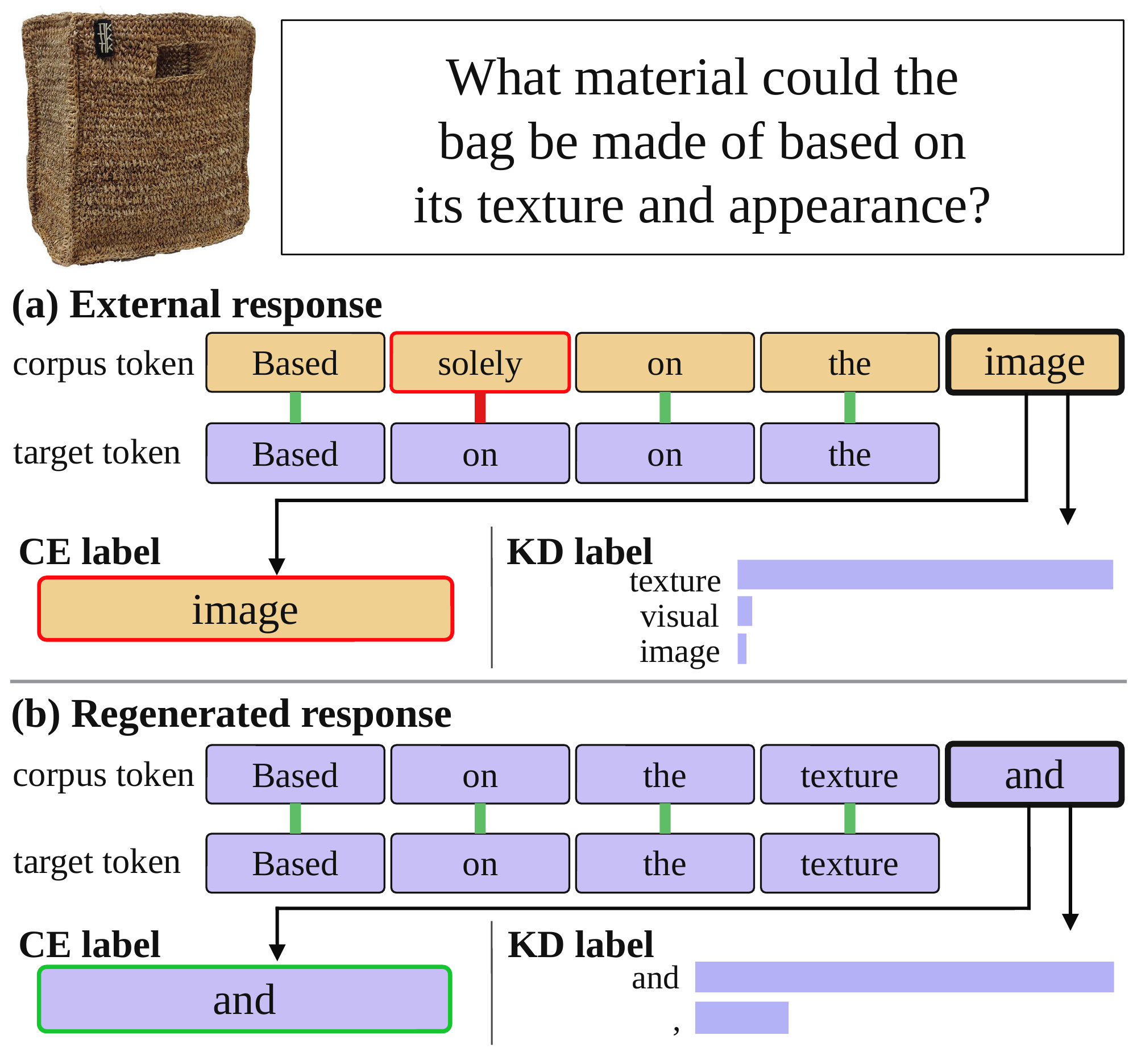}
\caption{One training example with external (a) and target-regenerated (b) responses. Below each corpus token is the target's greedy token, and red marks disagreement. Cross-entropy (CE) trains the outlined slot on its corpus token, distillation (KD) on the target's distribution.}
\label{fig:problem}
\end{figure}

Large language models (LLMs) and the vision-language models built on them \citep{yang2025qwen3,bai2025qwen3vl} generate text one token at a time \citep{kang2026thinkfuse,lee2026acoer,lee2026dart_routing}, and each token takes a forward pass through the full model \citep{lee2026bytecross}. Speculative decoding reduces the number of these sequential passes by letting a small draft model propose several tokens that the target model verifies in one parallel pass \citep{leviathan2023fast,chen2023accelerating}. This verification preserves the target's output distribution. The target accepts the drafted tokens until it rejects one, and the speed-up grows with the number of accepted tokens. Therefore, draft models are trained to predict what the target would generate. Recent drafters condition on the target's hidden states, predict several future tokens at once, or both \citep{cai2024medusa,li2025eagle3,an2025pard,chen2026dflash}. Beyond the architecture, other work designs the training objective, either distilling the target's token distributions into the drafter or giving more weight to positions that are more likely to survive verification \citep{zhou2024distillspec,an2026pard2,chen2026dflash}.

A new objective is often credited with the acceptance it adds over cross-entropy on a fixed corpus whose responses were written by other models \citep{li2024eagle,zhang2025hass}. However, many draft-training pipelines replace those responses with responses generated by the target itself \citep{cai2024medusa,goel2024direct,zafrir2025fastdraft,ganesan2025massv,chen2026dflash}. Such regeneration is sequence-level distillation from the target \citep{kim2016sequence}. As Figure~\ref{fig:problem} illustrates, the corpus token matches the target's own greedy prediction far more often in regenerated responses than in external ones. Where the two match, cross-entropy on the corpus already trains the drafter toward the target's prediction, and token-level distillation can add only the probability the target assigns to other tokens. Consequently, we expect part of an objective's gain to disappear after regeneration. If most of it disappears, a gain measured on external responses overstates what the objective adds in a pipeline that regenerates its responses.

To test this expectation, we measure how an objective's gain depends on who wrote the training responses. With Qwen3-VL-8B as the target and a DFlash-style block drafter, we compare cross-entropy, token-level distillation and the same distillation behind a gate that keeps only positions whose preceding corpus tokens would survive greedy verification. Each objective is trained on one image-grounded corpus with its original external responses and with responses the target regenerates greedily, and on a second corpus in the same two versions. Because the regenerated responses are also longer on average, we add two regenerated versions of the first corpus matched to the original supervised length, one that cuts each response and one that masks the loss beyond that length. Controls on the first corpus vary the initialization, the data order, the position weight, the form of the gate, the learning rate, the verifier and the decoding temperature. A horizon study varies the training length, and a text-only replication changes the target, the corpus and the verifier and adds an autoregressive drafter.

We find that the response source changes how much each objective appears to add. Distillation and the gate add large gains on external responses, but greedy regeneration removes most of both gains, leaving distillation a small gain and the gate little or none. The controls change the size of this shrinkage but not its direction. Most of the distillation shrinkage remains in both matched-length versions. Only the cut version leaves the gate a small gain, because cutting the responses instead of masking their loss lowers acceptance under the ungated objectives but not under the gated one. On external responses, cross-entropy loses more acceptance than distillation as training lengthens, so the distillation gain grows. At each objective's best measured training length, however, distillation adds much less on regenerated responses than on external ones. A second vision corpus shows the same shrinkage, and the text replication shows it for distillation and the gate under greedy and sampled decoding and for distillation with an autoregressive drafter under greedy decoding.

A label control supports the reasoning behind our expectation. On external responses, cross-entropy on the target's greedy tokens in place of the corpus tokens already yields most of the distillation gain. Our results suggest two practices for reporting objective gains. First, the gain of a new objective meant for pipelines that regenerate their responses should also be measured against cross-entropy on regenerated responses. Second, gains on external responses should be compared across training lengths.

\section{Study Design}
\label{sec:design}

\subsection{Drafter and objectives}

We study a parallel block drafter in the style of DFlash \citep{chen2026dflash}. At an anchor position $a$ of a training response $y$ to prompt $x$, the drafter reads hidden features of the frozen target up to $a$ and predicts the tokens at positions $a+1,\dots,a+15$ in one forward pass, one prediction for each slot $k=1,\dots,15$. During training, we sample up to 128 anchors in each example. An objective assigns each slot a loss $\ell_{a,k}$ and a gate $g_{a,k}\in\{0,1\}$. Following DFlash, slot $k$ is weighted by $w_k=\exp(-(k-1)/\gamma)$, and the loss is normalized by the retained weight:
\begin{equation}
\mathcal{L}=\frac{\sum_{a}\sum_{k} w_k\, g_{a,k}\, \ell_{a,k}}{\sum_{a}\sum_{k} w_k\, g_{a,k}} .
\label{eq:loss}
\end{equation}
Let $q_{a,k}$ be the drafter's distribution at slot $k$ and $p_{t}=p_T(\cdot\mid x,y_{<t})$ the target's next-token distribution at position $t$ under the corpus prefix. The three objectives differ in $\ell_{a,k}$ and $g_{a,k}$.

\paragraph{CE.} Cross-entropy on the corpus token, $\ell_{a,k}=-\log q_{a,k}(y_{a+k})$, with $g_{a,k}=1$. This is the DFlash objective applied to whatever responses the corpus contains.

\paragraph{KD.} Token-level distillation \citep{hinton2015distilling}, $\ell_{a,k}=-\sum_v \tilde p_{a+k}(v)\log \tilde q_{a,k}(v)$, with $g_{a,k}=1$. Here $\tilde p$ keeps the target's eight most probable tokens and merges the rest into a ninth outcome, and $\tilde q$ is the drafter's distribution over the same nine outcomes.

\paragraph{KD + gate.} KD behind a survival gate, which keeps only slots whose preceding corpus continuation would survive greedy verification,
\begin{equation}
g_{a,k}=\prod_{j=1}^{k-1}\mathbf{1}\!\left[y_{a+j}=\arg\max_v p_{a+j}(v)\right],
\end{equation}
with $g_{a,1}=1$. A slot that follows a corpus token other than the target's greedy token trains a prediction that greedy verification never uses, because the target does not produce that prefix. The gate is a hard counterpart of the survival weighting in PARD-2, which multiplies the target's confidence in the preceding corpus tokens \citep{an2026pard2}.

\subsection{Response treatments}

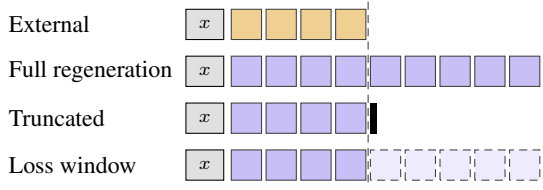
\begin{figure}[t]
\centering
\begin{tikzpicture}[x=1cm, y=1cm, font=\footnotesize]
  \def\bw{0.40}\def\bg{0.06}\def\bh{0.40}\def\xs{2.95}
  \newcommand{\tok}[4]{%
    \draw[#4, fill=#3, line width=0.4pt] ({\xs+#1*(\bw+\bg)}, {#2-\bh/2}) rectangle ++(\bw, \bh);}
  \foreach \y/\name in {3/External, 2/Full regeneration, 1/Truncated, 0/Loss window} {
    \node[anchor=west, inner sep=0pt] at (0, \y*0.62) {\name};
    \draw[fill=black!12, line width=0.4pt] (2.35, \y*0.62-\bh/2) rectangle ++(0.5, \bh);
    \node[inner sep=0pt] at (2.6, \y*0.62) {\scriptsize $x$};
  }
  \foreach \i in {0,...,3} {\tok{\i}{3*0.62}{extsand}{draw=black!75}}
  \foreach \i in {0,...,8} {\tok{\i}{2*0.62}{tgtlav}{draw=black!75}}
  \foreach \i in {0,...,3} {\tok{\i}{1*0.62}{tgtlav}{draw=black!75}}
  \fill[black] ({\xs+4*(\bw+\bg)}, {0.62-\bh/2}) rectangle ++(0.09, \bh);
  \foreach \i in {0,...,3} {\tok{\i}{0}{tgtlav}{draw=black!75}}
  \foreach \i in {4,...,8} {\tok{\i}{0}{tgtlav!30}{draw=black!75, densely dashed}}
  \draw[black!60, densely dashed, line width=0.5pt] ({\xs+4*(\bw+\bg)-\bg/2}, 3*0.62+0.30) -- ({\xs+4*(\bw+\bg)-\bg/2}, -0.30);
\end{tikzpicture}
\caption{Response treatments for one training example with image and prompt $x$. Sand and lavender blocks are external and target tokens, pale blocks carry no loss, the black bar closes the turn, and the dashed line marks the external length.}
\label{fig:treatments}
\end{figure}

All four treatments keep the images and prompts of the training corpus and change only the responses, as Figure~\ref{fig:treatments} illustrates for one example. They are (i) \emph{External}, the original corpus responses, written by a model other than the target; (ii) \emph{Full regeneration}, the target's greedy response to the same image and prompt, with at most 768 new tokens; (iii) \emph{Truncated}, the regenerated response cut to the token count of the external response, after which the assistant turn is closed; and (iv) \emph{Loss window}, the full regenerated response as input, with the loss restricted to the same number of leading response tokens as the external response.

Because regenerated responses are longer on average, both matched-length treatments supervise only as many response tokens as the external corpus. They differ at the cut, where truncation closes the turn and the loss window continues the response without supervision. The two therefore also differ in the anchors available near the cut and in batch padding. Neither isolates the effect of length, and we report both. For the analysis of Section~\ref{sec:analysis}, a fifth version replaces greedy regeneration with sampling at temperature 1, and a sixth samples at temperature 0.7, the default of the target's generation configuration.

\subsection{Measurement}

We report mean accepted tokens ($\MAT$), the number of tokens a verification round advances, counting the accepted draft tokens and the token the target adds itself, averaged over rounds. The increment of one objective over another is the ratio of their MATs minus one, for example $\MAT(\text{KD})/\MAT(\text{CE})-1$ on the same responses. To compare an increment across treatments, we use the ratio of ratios
\begin{equation}
\theta=\frac{\MAT_{r}(\text{KD})/\MAT_{r}(\text{CE})}{\MAT_{e}(\text{KD})/\MAT_{e}(\text{CE})}-1 ,
\end{equation}
where $r$ is a regenerated treatment and $e$ the external one. A negative $\theta$ means that KD adds less after regeneration. For the gate, KD + gate over KD replaces KD over CE.

\section{Experimental Setup}
\label{sec:setup}

\begin{table*}[t]
\centering
\fontsize{9}{10}\selectfont
\setlength{\tabcolsep}{3.8pt}
\renewcommand{\arraystretch}{0.92}
\setlength{\aboverulesep}{0.2ex}
\setlength{\belowrulesep}{0.4ex}
\begin{tabular*}{\textwidth}{@{\extracolsep{\fill}}lccccccc@{}}
\toprule
\rowcolor{ourblue!15}[0pt][0pt]\multicolumn{8}{@{}c@{}}{(a) Vision factorial\strut} \\
 & \multicolumn{2}{c}{Corpus (\%)} & \multicolumn{3}{c}{MAT} & \multicolumn{2}{c}{Increment (\%)} \\
\cmidrule(lr){2-3}\cmidrule(lr){4-6}\cmidrule(l){7-8}
Responses & Agreement & Kept & CE & KD & KD + gate & KD over CE & Gate over KD \\
\midrule
External & 58.0 & 17.3 & 2.875$_{\pm.009}$ & 3.291$_{\pm.008}$ & 3.453$_{\pm.007}$ & +14.4 [13.7, 15.2] & +4.9 [4.3, 5.6] \\
\rowcolor{groupyellow}[0pt][0pt]\multicolumn{8}{@{}c@{}}{\textit{Target regeneration}\strut} \\
Full & 95.2 & 73.4 & 3.843$_{\pm.030}$ & 3.987$_{\pm.002}$ & 3.965$_{\pm.012}$ & +3.7 [3.0, 4.5] & $-$0.5 [$-$1.1, 0.0] \\
Truncated & 94.4 & 73.3 & 3.483$_{\pm.014}$ & 3.724$_{\pm.008}$ & 3.807$_{\pm.031}$ & +6.9 [6.2, 7.7] & +2.2 [1.6, 2.9] \\
Loss window & 95.1 & 74.0 & 3.568$_{\pm.001}$ & 3.789$_{\pm.008}$ & 3.797$_{\pm.006}$ & +6.2 [5.4, 6.9] & +0.2 [$-$0.4, 0.8] \\
Sampled, $T=1$ & 76.4 & 31.3 & 3.756$_{\pm.043}$ & 3.915$_{\pm.009}$ & 3.997$_{\pm.035}$ & +4.2 [3.6, 4.8] & +2.1 [1.5, 2.7] \\
Sampled, $T=0.7$ & 84.1 & 41.5 & 3.810$_{\pm.020}$ & 3.947$_{\pm.015}$ & 3.989$_{\pm<.001}$ & +3.6 [2.9, 4.3] & +1.1 [0.5, 1.7] \\
\rowcolor{groupyellow}[0pt][0pt]\multicolumn{8}{@{}c@{}}{\textit{ShareGPT4V}\strut} \\
External & 64.7 & 19.4 & 2.802$_{\pm.003}$ & 3.307$_{\pm.005}$ & 3.507$_{\pm.015}$ & +18.0 [17.2, 18.8] & +6.1 [5.3, 6.8] \\
Full & 94.9 & 72.2 & 3.872$_{\pm.024}$ & 4.013$_{\pm.006}$ & 4.051$_{\pm.018}$ & +3.6 [3.0, 4.3] & +0.9 [0.3, 1.5] \\
\midrule
\rowcolor{ourblue!15}[0pt][0pt]\multicolumn{8}{@{}c@{}}{(b) Controls\strut} \\
Control & \multicolumn{2}{l}{Responses} & CE & KD & KD + gate & KD over CE & Gate over KD \\
\midrule
$\gamma=2$ & \multicolumn{2}{l}{External} & 2.890$_{\pm.010}$ & 3.365$_{\pm.020}$ & 3.526$_{\pm.006}$ & +16.4 [15.6, 17.3] & +4.8 [4.0, 5.6] \\
 & \multicolumn{2}{l}{Full} & 3.864$_{\pm.005}$ & 4.033$_{\pm.004}$ & 4.032$_{\pm.003}$ & +4.4 [3.6, 5.1] & 0.0 [$-$0.7, 0.6] \\
\addlinespace[3pt]
Chain verifier & \multicolumn{2}{l}{External} & 2.040$_{\pm.001}$ & 2.261$_{\pm<.001}$ & 2.397$_{\pm.005}$ & +10.8 [10.1, 11.5] & +6.0 [5.3, 6.7] \\
 & \multicolumn{2}{l}{Full} & 2.640$_{\pm.008}$ & 2.666$_{\pm.004}$ & 2.687$_{\pm.011}$ & +1.0 [0.4, 1.5] & +0.8 [0.3, 1.3] \\
 & \multicolumn{2}{l}{Truncated} & 2.448$_{\pm.018}$ & 2.540$_{\pm.005}$ & 2.569$_{\pm.002}$ & +3.8 [3.2, 4.3] & +1.2 [0.6, 1.7] \\
 & \multicolumn{2}{l}{Loss window} & 2.498$_{\pm.011}$ & 2.568$_{\pm.003}$ & 2.575$_{\pm.006}$ & +2.8 [2.3, 3.3] & +0.3 [$-$0.2, 0.8] \\
\addlinespace[3pt]
$T=1$ & \multicolumn{2}{l}{External} & 2.793$_{\pm.001}$ & 3.180$_{\pm.010}$ & 3.314$_{\pm.002}$ & +13.9 [13.0, 14.7] & +4.2 [3.5, 4.9] \\
 & \multicolumn{2}{l}{Full} & 3.576$_{\pm.012}$ & 3.738$_{\pm.007}$ & 3.740$_{\pm.014}$ & +4.5 [3.8, 5.2] & 0.0 [$-$0.6, 0.7] \\
\bottomrule
\end{tabular*}

\caption{Vision factorial and three ALLaVA controls after three epochs. Agreement is the share of slots whose corpus token is the target's greedy token, and kept the share the gate keeps. Subscripts give standard deviations over runs, brackets 95\% bootstrap intervals.}
\label{tab:main}
\end{table*}

\paragraph{Vision study.} The target is Qwen3-VL-8B-Instruct \citep{bai2025qwen3vl}. Apart from one control, every run continues training the released DFlash drafter for Qwen3-8B, whose language model has the same dimensions as the target's decoder \citep{chen2026dflash}. The training corpus is a fixed subset of 26,196 single-turn examples from ALLaVA \citep{chen2024allava}, and a subset of the same size from ShareGPT4V \citep{chen2024sharegpt4v} serves as a second corpus. The factorial trains for three epochs with $\gamma=7$ and the same learning rate for every objective and version. Each cell is trained twice, with different anchor sampling and the same data order, and every reported MAT averages the two runs. Appendix~\ref{app:vision-training} gives the remaining training details.

\paragraph{Evaluation.} Acceptance is measured on 99 scored prompts in each of three domains: detailed captions of COCO images \citep{lin2014coco}, TextVQA questions \citep{singh2019textvqa} from its training split, on which no drafter is trained, and DocVQA questions \citep{mathew2021docvqa}. The target decodes greedily for up to 256 tokens, and a tree verifier checks up to 63 drafted tokens in each round. A domain's MAT pools rounds over its prompts, and a run's MAT is the geometric mean of its three domain MATs. Intervals are 95\% percentile intervals from 10,000 resamples of prompts within each domain, so they describe prompt variation for the trained checkpoints.

\paragraph{Controls.} Eight controls test the vision results on external and fully regenerated ALLaVA responses. Six change training: random initialization; $\gamma=2$, with the soft survival weighting of PARD-2 as an extra objective; two further data orders; one third of the learning rate; an L2 anchor toward the released drafter; and KD over the target's 64 most probable tokens. Two change evaluation: a chain verifier that checks only the drafter's top candidate at each of seven slots, also on the matched-length versions, and target sampling at temperature 1 with random numbers shared by all drafters. Appendix~\ref{app:vision-controls} gives the objectives, epochs and runs of each control.

\paragraph{Text replication.} The target is Qwen3-4B with thinking disabled \citep{yang2025qwen3}, and the drafter continues from the released DFlash drafter for that model. The corpus pairs 24,552 first-turn prompts from UltraChat \citep{ding2023ultrachat} with either their original responses or the target's greedy responses, and CE, KD and KD + gate are trained on each version with the vision recipe. Evaluation uses the public DFlash decoder, which verifies one 16-token block in each round, on seven chat, math and code tasks, 2,103 prompts in total, at temperatures 0 and 1. For a second drafter family, a released drafter with the EAGLE-3 architecture \citep{li2025eagle3} for this target continues training with CE and KD on both versions and is evaluated at temperature 0 with its public tree decoder. Appendix~\ref{app:protocol} gives the remaining details.

\section{Results}
\label{sec:results}

\subsection{Regeneration shrinks objective gains}

Table~\ref{tab:main}a reports the vision factorial. On ALLaVA, distillation and the gate add large gains on external responses, but after full regeneration KD adds 3.7\% over CE and the gate no measurable gain. The ratio of ratios is $-$9.4\% for KD and $-$5.2\% for the gate, and both are negative in every domain and training run. Regeneration raises MAT under every objective, CE the most and KD + gate the least, so the three objectives end close together, and the objective that gained most on external responses gains least from the new corpus. ShareGPT4V shows the same shrinkage: after full regeneration KD's gain over CE falls from 18.0\% to 3.6\% and the gate's gain over KD from 6.1\% to 0.9\%, and the ratios of ratios, $-$12.2\% for KD and $-$4.8\% for the gate, are negative in every domain and training run.

\begin{figure}[t]
\centering
\includegraphics{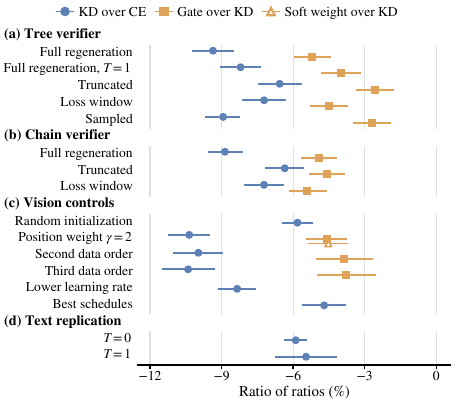}
\caption{Ratios of ratios with 95\% prompt bootstrap intervals, comparing an increment on the given responses with the same increment on external responses. Data-order rows come from one training run, and best schedules use each objective's best measured training length.}
\label{fig:theta}
\end{figure}

\begin{table}[t]
\centering
{
\fontsize{7}{8.2}\selectfont
\renewcommand{\arraystretch}{0.96}
\setlength{\tabcolsep}{0.8pt}
\setlength{\aboverulesep}{0.2ex}\setlength{\belowrulesep}{0.4ex}
\begin{tabular*}{\linewidth}{@{\extracolsep{\fill}}llccc@{}}
\toprule
Condition & Responses & CE & KD & KD gain (\%) \\
\midrule
\rowcolor{groupyellow}[0pt][0pt]\multicolumn{5}{@{}c@{}}{\textit{Vision controls}\strut} \\
\multirow[t]{2}{*}{Rand. init., 3 ep.} & External & 2.066$_{\pm.002}$ & 2.259$_{\pm.002}$ & +9.3 [8.7, 9.9] \\
 & Full & 2.828$_{\pm.002}$ & 2.912$_{\pm.029}$ & +3.0 [2.4, 3.5] \\
\cmidrule(l){2-5}
\multirow[t]{2}{*}{Rand. init., 6 ep.} & External & 2.071$_{\pm.022}$ & 2.349$_{\pm.020}$ & +13.4 [12.7, 14.2] \\
 & Full & 3.082$_{\pm.014}$ & 3.200$_{\pm.041}$ & +3.8 [3.2, 4.4] \\
\cmidrule(l){2-5}
\multirow[t]{2}{*}{Lower LR} & External & 3.099$_{\pm.007}$ & 3.474$_{\pm.016}$ & +12.1 [11.3, 12.9] \\
 & Full & 3.893$_{\pm.009}$ & 4.000$_{\pm.002}$ & +2.8 [2.2, 3.3] \\
\cmidrule(l){2-5}
\multirow[t]{2}{*}{L2 anchor} & External & 3.152$_{\pm.007}$ & 3.474$_{\pm<.001}$ & +10.2 [9.4, 11.0] \\
 & Full & 3.937$_{\pm.013}$ & 4.005$_{\pm.014}$ & +1.7 [1.1, 2.3] \\
\rowcolor{groupyellow}[0pt][0pt]\multicolumn{5}{@{}c@{}}{\textit{Text replication}\strut} \\
\multirow[t]{2}{*}{DFlash, $T=0$} & External & 3.227$_{\pm.008}$ & 3.464$_{\pm.004}$ & +7.3 [6.8, 7.8] \\
 & Regenerated & 4.141$_{\pm.016}$ & 4.183$_{\pm.006}$ & +1.0 [0.7, 1.3] \\
\cmidrule(l){2-5}
\multirow[t]{2}{*}{DFlash, $T=1$} & External & 3.018$_{\pm.004}$ & 3.225$_{\pm.017}$ & +6.9 [6.0, 7.7] \\
 & Regenerated & 3.750$_{\pm.022}$ & 3.789$_{\pm.009}$ & +1.0 [0.0, 2.0] \\
\cmidrule(l){2-5}
\multirow[t]{2}{*}{EAGLE-3, $T=0$} & External & 3.886$_{\pm.017}$ & 4.420$_{\pm.020}$ & +13.7 [13.2, 14.3] \\
 & Regenerated & 4.791$_{\pm.009}$ & 4.895$_{\pm.002}$ & +2.2 [1.8, 2.5] \\
\midrule
Condition & Responses & KD & KD + gate & Gate gain (\%) \\
\midrule
\rowcolor{groupyellow}[0pt][0pt]\multicolumn{5}{@{}c@{}}{\textit{Text replication: gate}\strut} \\
\multirow[t]{2}{*}{DFlash, $T=0$} & External & 3.464$_{\pm.004}$ & 3.623$_{\pm.002}$ & +4.6 [4.1, 5.0] \\
 & Regenerated & 4.183$_{\pm.006}$ & 4.188$_{\pm.005}$ & +0.1 [$-$0.2, 0.4] \\
\cmidrule(l){2-5}
\multirow[t]{2}{*}{DFlash, $T=1$} & External & 3.225$_{\pm.017}$ & 3.362$_{\pm.004}$ & +4.2 [3.2, 5.2] \\
 & Regenerated & 3.789$_{\pm.009}$ & 3.783$_{\pm.001}$ & $-$0.2 [$-$1.2, 0.9] \\
\bottomrule
\end{tabular*}\par

}
\caption{MAT after three epochs unless noted. Subscripts show run standard deviations, brackets 95\% bootstrap intervals. Gains compare KD with CE and KD + gate with KD.}
\label{tab:cekd}
\end{table}

Table~\ref{tab:main}b repeats the factorial comparison under three controls that cover all three objectives. In each of them KD adds more than 10\% and the gate more than 4\% on external responses, whereas after full regeneration KD adds less than 5\% and the gate less than 1\%. Figure~\ref{fig:theta} shows that every control leaves the shrinkage in place. Like the gate, the soft survival weighting of PARD-2 adds little after regeneration. Under the chain verifier the ratios of ratios stay close to those of the tree verifier, and when the target samples at temperature 1 they are $-$8.2\% for KD and $-$4.0\% for the gate.

Table~\ref{tab:cekd} reports the three vision controls that train only CE and KD. From random initialization, regenerated CE reaches 2.828 after three epochs, close to the 2.875 of external CE from the released drafter, yet KD adds 3.0\% over CE there against 9.3\% on external responses, so a higher baseline does not explain the smaller increment. Training this control twice as long moves its ratio of ratios from $-$5.8\% to $-$8.5\%. At one third of the learning rate, KD adds 12.1\% over CE on external responses, against 14.4\% in the factorial, and its ratio of ratios is $-$8.3\%. With the L2 anchor on both objectives, KD adds 10.2\% over CE on external and 1.7\% on regenerated responses, a ratio of ratios of $-$7.7\% against $-$9.4\% without the anchor, so most of the shrinkage remains. Appendix~\ref{app:controls} gives every ratio of ratios, the cells with the soft weighting and the data orders.

\subsection{Shrinkage persists at matched length}

Table~\ref{tab:main}a also shows that most of the shrinkage remains when the supervised length is matched. On ALLaVA, regenerated responses are 3.5 times as long as external ones on average, so full regeneration also adds supervised positions and anchors. After truncation and with the loss window, KD adds 6.9\% and 6.2\% over CE, less than half of its external increment, and both ratios of ratios are negative in every domain and run.

In Table~\ref{tab:main}a the two length controls disagree about the gate, which adds 2.2\% under truncation and 0.2\% under the loss window, a ratio of ratios between them of 2.0\%. Relative to the loss window, truncation lowers CE by 2.4\% and KD by 1.7\% but leaves KD + gate within 0.3\%, so the gate recovers what truncation takes from the ungated objectives. A length control built by cutting responses can therefore show a gate gain that the loss window does not. Masking the turn-closing tokens that the cut leaves as training targets removes 80\% of the gap between the gate's gains under the two treatments and leaves no measurable gain beyond the loss window's, so turn closure explains most of the gate's truncation-only gain. Under the chain verifier, the two gate ratios of ratios in Figure~\ref{fig:theta} lie within one point of each other.

\begin{figure}[t]
\centering
\includegraphics{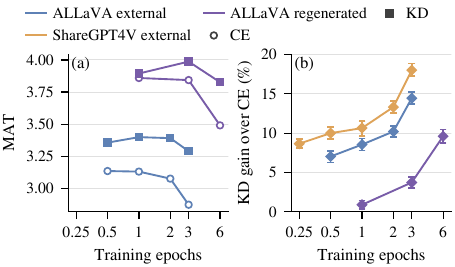}
\caption{(a) Vision MAT of CE and KD on ALLaVA by training length. (b) KD gain over CE with 95\% prompt bootstrap intervals. Every point is a separately trained and annealed schedule, not a checkpoint of a longer run.}
\label{fig:horizon}
\end{figure}

\subsection{Training length moves objective gains}

Figure~\ref{fig:horizon} varies the length of training. On external responses the KD gain grows with training beyond half an epoch on both corpora, because CE degrades with longer training while KD changes less. On ALLaVA, CE loses 8.2\% of its one-epoch MAT by the third epoch and KD 3.2\%. On regenerated ALLaVA responses the gain grows from 0.9\% after one epoch to 3.7\% after three because KD improves while CE stays within 1.4\% of its two-epoch MAT. It reaches 5.6\% after four epochs and 9.6\% after six because both objectives then degrade and CE degrades more, as on external responses. A comparison at a single training length can therefore overstate or understate an objective's gain. Regenerated ALLaVA KD peaks at three epochs, inside the measured range. At each objective's best measured schedule, KD adds 8.4\% over CE on external and 2.4\% on regenerated ALLaVA responses, a ratio of ratios of $-$5.6\%. This is about three fifths of the three-epoch shrinkage and remains well below zero. For the gate, the same comparison gives 2.7\% over KD on external and $-$0.5\% on regenerated ALLaVA responses, a ratio of ratios of $-$3.1\%. Table~\ref{tab:horizon} lists every CE and KD schedule, and Table~\ref{tab:app-horizon-gate} the gate schedules.

\begin{table}[t]
\centering
\small
\renewcommand{\arraystretch}{0.92}
\setlength{\aboverulesep}{0.2ex}
\setlength{\belowrulesep}{0.4ex}
\fontsize{8}{9.3}\selectfont
\renewcommand{\arraystretch}{0.95}
\setlength{\aboverulesep}{0.2ex}\setlength{\belowrulesep}{0.4ex}\setlength{\fboxsep}{0pt}
\setlength{\tabcolsep}{1.5pt}
\begin{tabular*}{\linewidth}{@{\extracolsep{\fill}}llcccc@{}}
\toprule
Corpus & Responses & Epochs & CE & KD & Gain (\%) \\
\midrule
\multirow[t]{10}{*}{ALLaVA} & \multirow[t]{5}{*}{External} & 0.25 & 3.098 & 3.318 & +7.1 \\
 &  & 0.5 & 3.137 & 3.357 & +7.0 \\
 &  & 1 & 3.132 & 3.400 & +8.5 \\
 &  & 2 & 3.078 & 3.392 & +10.2 \\
 &  & 3 & 2.875 & 3.291 & +14.4 \\
\cmidrule(l){2-6}
 & \multirow[t]{5}{*}{Regenerated} & 1 & 3.859 & 3.894 & +0.9 \\
 &  & 2 & 3.895 & 3.969 & +1.9 \\
 &  & 3 & 3.843 & 3.987 & +3.7 \\
 &  & 4 & 3.718 & 3.927 & +5.6 \\
 &  & 6 & 3.491 & 3.827 & +9.6 \\
\midrule
\multirow[t]{9}{*}{ShareGPT4V} & \multirow[t]{5}{*}{External} & 0.25 & 3.140 & 3.412 & +8.7 \\
 &  & 0.5 & 3.144 & 3.459 & +10.0 \\
 &  & 1 & 3.104 & 3.434 & +10.6 \\
 &  & 2 & 3.003 & 3.402 & +13.3 \\
 &  & 3 & 2.802 & 3.307 & +18.0 \\
\cmidrule(l){2-6}
 & \multirow[t]{4}{*}{Regenerated} & 0.5 & 3.895 & 3.891 & $-$0.1 \\
 &  & 1 & 3.984 & 4.012 & +0.7 \\
 &  & 3 & 3.872 & 4.013 & +3.6 \\
 &  & 6 & 3.499 & 3.838 & +9.7 \\
\bottomrule
\end{tabular*}

\caption{Vision MAT and KD gains over CE across corpora, response sources and training lengths. Each schedule is trained and annealed separately.}
\label{tab:horizon}
\end{table}

On ALLaVA, after one epoch the KD ratio of ratios is $-$7.0\% and is negative in every domain and run. The best schedules are those with the highest MAT on the evaluation prompts. On external ALLaVA responses they are half an epoch for CE and one epoch for KD, and on regenerated ones two epochs for CE and three for KD. Training external ALLaVA CE for a quarter epoch lowers its MAT by 1.2\% relative to half an epoch, so half an epoch remains its best measured schedule. When the schedules are instead selected on alternate prompts of each domain and the ratio of ratios is estimated on the remaining prompts, both halves select the same ALLaVA schedules, and the KD best-schedule ratio of ratios averaged over the two directions is $-$5.6\%.

From three to six epochs, regenerated ALLaVA MAT falls by 0.35 for CE and by 0.16 for KD. Regenerated ALLaVA CE is best at two epochs, so both regenerated optima lie inside the measured range. Restricted to the one-, two- and three-epoch schedules that both sources share, the KD best-schedule ratio of ratios is $-$5.7\%. On ShareGPT4V, comparing each objective at its best measured schedule leaves KD 10.0\% ahead of CE on external and 0.7\% on regenerated responses, a ratio of ratios of $-$8.4\%.

\subsection{A text-only replication}

Table~\ref{tab:cekd} and Figure~\ref{fig:theta} also include the text-only replication. KD adds 7.3\% over CE on external and 1.0\% on regenerated responses under greedy decoding, a ratio of ratios of $-$5.9\%, and the ratio of ratios under sampling is $-$5.5\%. The ratio of ratios is negative in all seven tasks under greedy decoding and in six under sampling, the exception being AIME 2025 with 30 prompts at $+$0.6\%. The gate adds 4.6\% over KD on external responses and no measurable gain on regenerated ones under greedy decoding, a ratio of ratios of $-$4.3\%, and its ratio of ratios under sampling is $-$4.2\%. Appendix~\ref{app:text} reports the results for each task.

Table~\ref{tab:cekd} also lists an autoregressive EAGLE-3 drafter trained with CE and KD on the same two versions. KD adds 13.7\% over CE on external and 2.2\% on regenerated responses under greedy decoding, a ratio of ratios of $-$10.2\% that is negative in all seven tasks and in both training runs.

\section{Why the Increments Shrink}
\label{sec:analysis}

\paragraph{The corpus already carries the target's greedy token.} The corpus columns of Table~\ref{tab:main}a measure how often the corpus token equals the target's greedy token. On ALLaVA, this agreement rises from 58.0\% on external responses to 95.2\% after full regeneration, and on ShareGPT4V from 64.7\% to 94.9\%. With greedily regenerated responses, CE already trains toward the target's greedy prediction at nearly every slot, KD can add only the probability mass away from that token, and the gate has few continuations left to remove. Agreement separates the external versions from the target-written ones, but it does not order the versions within each group by their KD gains. Training CE and KD on three target samples of every prompt at temperature 1 for the same number of updates removes most of KD's gain on sampled responses, so more target samples also supply most of that gain. Appendix~\ref{app:corpus} defines agreement exactly and compares it across versions, and Appendix~\ref{app:controls} reports the three-sample cells.

\begin{figure}[t]
\centering
\includegraphics{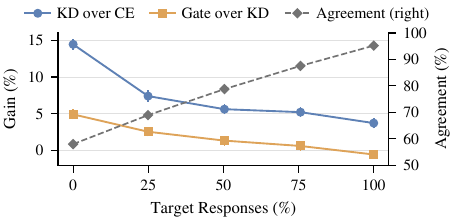}
\caption{Objective gains when a share of the external ALLaVA responses is replaced by the target's greedy regenerations, with 95\% intervals.}
\label{fig:dose}
\end{figure}

\paragraph{The gains fall with the share the target wrote.} We also train each objective, twice, on mixtures that replace a random quarter, half or three quarters of the external ALLaVA responses with their greedy regenerations, keeping the prompts, their order and the training length of the external cells. As Figure~\ref{fig:dose} shows, both gains decline as the regenerated share grows, while agreement rises. Most of the KD decline comes with the first quarter: regenerating a quarter of the responses already covers about two thirds of the drop from external to full regeneration, so a KD gain measured on such a mixture is closer to the fully regenerated gain than to the external one. Appendix~\ref{app:analysis} lists every mixture.

\paragraph{Greedy labels carry the part of the external gain that regeneration removes.} A label control on ALLaVA separates two parts of what KD adds. Training on the target's greedy token in place of the corpus token at every position, with the corpus prefix unchanged, raises MAT by 9.3\% over CE on external responses, and KD on the same prefixes adds a further 4.7\%. On fully regenerated responses the same control adds no measurable gain over CE, and KD adds 3.9\% over it, close to the 3.7\% that KD keeps over CE. In the vision study, regeneration thus removes the greedy-label part of the external KD gain and leaves most of what KD adds beyond the target's greedy token. In the text replication under greedy decoding, the same control matches KD on external responses and adds only 0.3\% over CE on regenerated ones. Appendix~\ref{app:vision} lists the control for each domain and Appendix~\ref{app:text} for each task.

\paragraph{The remaining KD gain lies beyond the drafter's first choice.} On external ALLaVA responses KD mostly corrects the drafter's first choice, whereas after greedy regeneration most of its gain lies in how the other tree candidates are ranked. Appendix~\ref{app:analysis} shows this by scoring each drafter on the target's greedy text.

\paragraph{Full regeneration outweighs the objective settings we tried.} From the released drafter's 2.835, three epochs of CE on external ALLaVA responses end 1.4\% higher, whereas every regenerated cell of Table~\ref{tab:main}a ends at least 22.9\% higher. Fully regenerated CE also exceeds every objective setting we trained on external ALLaVA responses. It is 8.5\% above the best of these, KD with the soft survival weighting of PARD-2 and $\gamma=2$. Table~\ref{tab:cekd} and Appendix~\ref{app:settings} list the settings.

\section{Related Work}
\label{sec:related}

\paragraph{Drafters and objectives.} Work on speculative decoding \citep{leviathan2023fast,chen2023accelerating} has proposed both drafter architectures and training objectives. Autoregressive drafters predict one token at a time \citep{li2024eagle2,li2024eagle,li2025eagle3}, and parallel drafters predict several future tokens in one forward pass \citep{cai2024medusa,an2025pard,chen2026dflash,lee2026cast,lee2026redline}. Distillation objectives train the drafter to match the target's token distributions \citep{liu2023online,zhou2024distillspec,goel2024direct,zhang2025hass}. Other objectives give less weight to the positions that follow a corpus token the target or the drafter is unlikely to produce. PARD-2 weights tokens by the cumulative target confidence of the preceding corpus tokens \citep{an2026pard2}, and GRIFFIN masks the loss after the first corpus token outside the drafter's own top-$k$ predictions \citep{hu2025griffin}. Our gate likewise masks the loss after the first corpus token that is not the target's greedy token, so it is a hard counterpart of the PARD-2 weighting. We also test the PARD-2 weighting itself in the control with $\gamma=2$.

\paragraph{Who writes the training responses.} Training drafters on target-generated responses is established practice \citep{cai2024medusa,goel2024direct,zafrir2025fastdraft,ganesan2025massv}, and DFlash, the drafter family used in both of our studies, regenerates its training responses with the target \citep{chen2026dflash}. On such responses, \citet{hong2025training} find that distilling the target's token distributions gives drafters for domain-specific targets a 2\% to 10\% higher acceptance rate than cross-entropy on the same responses. Other drafters are trained and compared on fixed responses. EAGLE uses a fixed corpus to avoid the cost of generation \citep{li2024eagle}, and HASS compares its distillation loss with six alternative losses on that corpus \citep{zhang2025hass}. Two concurrent works propose a draft-training method that relabels an external corpus with greedy target rollouts \citep{lee2026recovering} and a vision-language block drafter trained on the target's greedy generations \citep{lee2026glance}. We propose neither and instead treat the response source as an experimental factor, train the same objectives on both sources, and measure how regeneration changes what each objective adds. Student-generated data with target feedback \citep{agarwal2024policy,zhou2024distillspec} is a different factor, because it trains on the drafter's own outputs.

\paragraph{Distillation data beyond drafting.} Sequence-level distillation trains a student on teacher-generated sequences instead of matching token distributions, and in translation, adding word-level distillation on top of it changed BLEU by less than one point for every student \citep{kim2016sequence}. Distilled corpora differ in complexity in ways whose value depends on student capacity \citep{zhou2019understanding}. In language model distillation, how the generated training data are constructed can decide whether distillation helps \citep{lee2026answerconditioned,lee2026directional}, and the effect of a training objective can depend on the amount of available data \citep{lee2025clo}. Draft training on regenerated responses is sequence-level distillation from the target, with KD as the word-level term.

\section{Discussion}
\label{sec:discussion}

We find that the gain of a draft-training objective depends on who wrote the training responses. In our vision-language study, distillation and the survival gate add large gains on external responses of both corpora, but once the target regenerates them greedily, distillation adds a small gain and the gate little or none. Even at matched supervised length on ALLaVA or at each objective's best measured training length, distillation adds less than half of its external gain, and a text-only replication shows the shrinkage too, for distillation also with an autoregressive drafter. Because regeneration is already common in draft training, the external-corpus gain of a new objective can overstate what it adds in such a pipeline.

\paragraph{Which gains survive regeneration.} We expect objectives that repair a mismatch between corpus and target to lose most of their value in pipelines that regenerate. The gate and the soft survival weighting discount corpus continuations that the target would not produce, and greedy regeneration leaves few such continuations. KD instead adds the target's probability mass beyond its greedy token, which regenerated responses do not contain, and keeps a small gain. HASS, which adds a top-$K$ distillation loss and trains the drafter on its own predicted features, likewise keeps its gain over EAGLE-2 \citep{li2024eagle2} when both are trained on target-generated rather than fixed responses \citep{zhang2025hass}.

\paragraph{Reporting objective gains.} For an objective meant for pipelines that regenerate their responses, we suggest also measuring its gain over CE on responses that the target regenerates with the pipeline's own decoding. Regenerating the ALLaVA and text corpora cost less than one training run, as Appendix~\ref{app:protocol} shows. KD already computes the target's distribution at every corpus position, so the agreement between corpus tokens and the target's greedy tokens comes at no extra cost wherever KD is trained. Reporting this agreement beside an objective gain shows how far the training responses are from the target's own greedy predictions. Gains on external responses should be compared across training lengths or at each objective's best schedule, because the external KD gain roughly doubled from the shortest to the longest schedule we trained.

\section*{Limitations}

\paragraph{Drafters and targets.} We study two drafter families with two Qwen3 targets: a DFlash-style parallel block drafter in both studies and an autoregressive EAGLE-3 drafter in the text study, where it is trained only with CE and KD and evaluated only under greedy decoding. Other drafter families, the gate on an autoregressive drafter and targets from other families may respond differently. Apart from one control, every vision drafter continues training a released drafter. That control shows the distillation shrinkage from random initialization at a smaller magnitude after three epochs, so the magnitudes we report describe continued training. When the drafters of both objectives are anchored to the released drafter, most of the distillation shrinkage remains.

\paragraph{Corpora and training lengths.} Limited compute kept each training corpus to a fixed subset of at most 26,196 examples, and we did not extend the study to the full corpora. The second corpus, ShareGPT4V, has only external and fully regenerated responses, and the matched-length versions and the controls cover ALLaVA alone. The horizon study measures the gate only on ALLaVA. Its best schedules are selected on the evaluation prompts, although the best-schedule ratio of ratios stays below zero on both corpora when the schedules are chosen on one half of the prompts and the ratio is estimated on the other.

\paragraph{Measurement.} MAT depends on the verification decoder. We measure it with a tree and a chain verifier in the vision study and, in the text study, with a block verifier for the DFlash drafter and the EAGLE-3 tree verifier for the EAGLE-3 drafter, and we do not measure wall-clock speed or answer quality. The vision verifiers accept drafts against the target's greedy token computed within each batched bf16 verification pass, so MAT describes acceptance under that execution.

\paragraph{Runs and intervals.} Each DFlash cell has two training runs with one data order and each EAGLE-3 cell two runs in two data orders, six vision cells add one run in each of two further data orders, and each domain has 99 scored prompts. The bootstrap intervals cover prompt variation for the trained checkpoints, and the data orders vary only for those six cells and the EAGLE-3 cells. Appendix~\ref{app:vision} also resamples the two runs of each cell, and the full-regeneration intervals of both corpora stay below zero. Two runs give only a coarse account of run variation, and no interval covers variation across corpora.

\paragraph{Objectives and treatments.} The factorial uses a hard survival gate, one distillation divergence, $\gamma=7$ and one learning rate. The soft survival weighting and $\gamma=2$ are tested only on external and fully regenerated responses, and the lower learning rate only with CE and KD on those responses. Other divergences and position weights may shrink differently after regeneration. The response treatments change content, length, turn boundaries and anchor support together, so the matched-length controls bound the role of length without isolating it.

\section*{Ethical Considerations}

This work trains and evaluates draft models with public datasets and released models within their licenses. It involves no human subjects and collects no data from people. A draft model changes how quickly its target generates, not what the target can produce, so the behavior and risks of the target carry over unchanged. Appendix~\ref{app:licenses} lists the licenses of all artifacts.

\bibliography{references_arxiv}

\begin{thebibliography}{47}
\providecommand{\natexlab}[1]{#1}

\bibitem[{Agarwal et~al.(2024)Agarwal, Vieillard, Zhou, Stanczyk, Ramos~Garea,
  Geist, and Bachem}]{agarwal2024policy}
Rishabh Agarwal, Nino Vieillard, Yongchao Zhou, Piotr Stanczyk, Sabela
  Ramos~Garea, Matthieu Geist, and Olivier Bachem. 2024.
\newblock \href
  {https://proceedings.iclr.cc/paper_files/paper/2024/hash/5be69a584901a26c521c2b51e40a4c20-Abstract-Conference.html}
  {{On-Policy Distillation of Language Models: Learning from Self-Generated
  Mistakes}}.
\newblock In \emph{International Conference on Learning Representations}.

\bibitem[{An et~al.(2026{\natexlab{a}})An, Bai, Liu, Li, and
  Barsoum}]{an2025pard}
Zihao An, Huajun Bai, Ziqiong Liu, Dong Li, and Emad Barsoum.
  2026{\natexlab{a}}.
\newblock \href {https://openreview.net/forum?id=XbOyv7iVGL} {{PARD:
  Accelerating LLM Inference with Low-Cost PARallel Draft Model Adaptation}}.
\newblock In \emph{International Conference on Learning Representations}.

\bibitem[{An et~al.(2026{\natexlab{b}})An, Liu, Liu, Li, Liu, and
  Barsoum}]{an2026pard2}
Zihao An, Taichi Liu, Ziqiong Liu, Dong Li, Ruofeng Liu, and Emad Barsoum.
  2026{\natexlab{b}}.
\newblock \href {https://doi.org/10.48550/arxiv.2605.08632} {{PARD-2:
  Target-Aligned Parallel Draft Model for Dual-Mode Speculative Decoding}}.
\newblock \emph{arXiv preprint arXiv:2605.08632}.

\bibitem[{Bai et~al.(2025)Bai, Cai, Chen, Chen, Chen, Cheng, Deng, Ding, Gao,
  Ge, Ge, Guo, Huang, Huang, Huang, Hui, Jiang, Li, Li, Li, Li, Lin, Lin, Liu,
  Liu, Liu, Liu, Liu, Liu, Lu, Luo, Lv, Men, Meng, Ren, Ren, Song, Sun, Tang,
  Tu, Wan, Wang, Wang, Wang, Wang, Xie, Xu, Xu, Xu, Yang, Yang, Yang, Yang, Yu,
  Zhang, Zhang, Zhang, Zheng, Zhong, Zhou, Zhou, Zhou, Zhu, and
  Zhu}]{bai2025qwen3vl}
Shuai Bai, Yuxuan Cai, Ruizhe Chen, Keqin Chen, Xionghui Chen, Zesen Cheng,
  Lianghao Deng, Wei Ding, Chang Gao, Chunjiang Ge, Wenbin Ge, Zhifang Guo,
  Qidong Huang, Jie Huang, Fei Huang, Binyuan Hui, Shutong Jiang, Zhaohai Li,
  Mingsheng Li, and 45 others. 2025.
\newblock \href {https://doi.org/10.48550/arxiv.2511.21631} {{Qwen3-VL
  Technical Report}}.
\newblock \emph{arXiv preprint arXiv:2511.21631}.

\bibitem[{Cai et~al.(2024)Cai, Li, Geng, Peng, Lee, Chen, and
  Dao}]{cai2024medusa}
Tianle Cai, Yuhong Li, Zhengyang Geng, Hongwu Peng, Jason~D. Lee, Deming Chen,
  and Tri Dao. 2024.
\newblock \href {https://proceedings.mlr.press/v235/cai24b.html} {{Medusa:
  Simple LLM Inference Acceleration Framework with Multiple Decoding Heads}}.
\newblock In \emph{International Conference on Machine Learning}.

\bibitem[{Chen et~al.(2023)Chen, Borgeaud, Irving, Lespiau, Sifre, and
  Jumper}]{chen2023accelerating}
Charlie Chen, Sebastian Borgeaud, Geoffrey Irving, Jean-Baptiste Lespiau,
  Laurent Sifre, and John Jumper. 2023.
\newblock \href {https://doi.org/10.48550/arxiv.2302.01318} {{Accelerating
  Large Language Model Decoding with Speculative Sampling}}.
\newblock \emph{arXiv preprint arXiv:2302.01318}.

\bibitem[{Chen et~al.(2024{\natexlab{a}})Chen, Chen, Zhang, Chen, Wu, Zhang,
  Chen, Li, Wan, and Wang}]{chen2024allava}
Guiming~Hardy Chen, Shunian Chen, Ruifei Zhang, Junying Chen, Xiangbo Wu, Zhiyi
  Zhang, Zhihong Chen, Jianquan Li, Xiang Wan, and Benyou Wang.
  2024{\natexlab{a}}.
\newblock \href {https://doi.org/10.48550/arxiv.2402.11684} {{ALLaVA:
  Harnessing GPT4V-Synthesized Data for Lite Vision-Language Models}}.
\newblock \emph{arXiv preprint arXiv:2402.11684}.

\bibitem[{Chen et~al.(2026)Chen, Liang, and Liu}]{chen2026dflash}
Jian Chen, Yesheng Liang, and Zhijian Liu. 2026.
\newblock \href {https://openreview.net/forum?id=Oz335dV48X} {{DFlash: Block
  Diffusion for Flash Speculative Decoding}}.
\newblock In \emph{International Conference on Machine Learning}.

\bibitem[{Chen et~al.(2024{\natexlab{b}})Chen, Li, Dong, Zhang, He, Wang, Zhao,
  and Lin}]{chen2024sharegpt4v}
Lin Chen, Jinsong Li, Xiaoyi Dong, Pan Zhang, Conghui He, Jiaqi Wang, Feng
  Zhao, and Dahua Lin. 2024{\natexlab{b}}.
\newblock \href {https://doi.org/10.1007/978-3-031-72643-9_22} {{ShareGPT4V:
  Improving Large Multi-modal Models with Better Captions}}.
\newblock In \emph{Computer Vision -- ECCV 2024}, volume 15075 of \emph{Lecture
  Notes in Computer Science}, pages 370--387. Springer.

\bibitem[{Chen et~al.(2021)Chen, Tworek, Jun, Yuan, de~Oliveira~Pinto, Kaplan,
  Edwards, Burda, Joseph, Brockman, Ray, Puri, Krueger, Petrov, Khlaaf, Sastry,
  Mishkin, Chan, Gray, Ryder, Pavlov, Power, Kaiser, Bavarian, Winter, Tillet,
  Such, Cummings, Plappert, Chantzis, Barnes, Herbert-Voss, Guss, Nichol,
  Paino, Tezak, Tang, Babuschkin, Balaji, Jain, Saunders, Hesse, Carr, Leike,
  Achiam, Misra, Morikawa, Radford, Knight, Brundage, Murati, Mayer, Welinder,
  McGrew, Amodei, McCandlish, Sutskever, and Zaremba}]{chen2021evaluating}
Mark Chen, Jerry Tworek, Heewoo Jun, Qiming Yuan, Henrique~Ponde
  de~Oliveira~Pinto, Jared Kaplan, Harri Edwards, Yuri Burda, Nicholas Joseph,
  Greg Brockman, Alex Ray, Raul Puri, Gretchen Krueger, Michael Petrov, Heidy
  Khlaaf, Girish Sastry, Pamela Mishkin, Brooke Chan, Scott Gray, and 39
  others. 2021.
\newblock \href {https://arxiv.org/abs/2107.03374} {{Evaluating Large Language
  Models Trained on Code}}.
\newblock \emph{arXiv preprint arXiv:2107.03374}.

\bibitem[{Cobbe et~al.(2021)Cobbe, Kosaraju, Bavarian, Chen, Jun, Kaiser,
  Plappert, Tworek, Hilton, Nakano, Hesse, and Schulman}]{cobbe2021training}
Karl Cobbe, Vineet Kosaraju, Mohammad Bavarian, Mark Chen, Heewoo Jun, Lukasz
  Kaiser, Matthias Plappert, Jerry Tworek, Jacob Hilton, Reiichiro Nakano,
  Christopher Hesse, and John Schulman. 2021.
\newblock \href {https://arxiv.org/abs/2110.14168} {{Training Verifiers to
  Solve Math Word Problems}}.
\newblock \emph{arXiv preprint arXiv:2110.14168}.

\bibitem[{Ding et~al.(2023)Ding, Chen, Xu, Qin, Hu, Liu, Sun, and
  Zhou}]{ding2023ultrachat}
Ning Ding, Yulin Chen, Bokai Xu, Yujia Qin, Shengding Hu, Zhiyuan Liu, Maosong
  Sun, and Bowen Zhou. 2023.
\newblock \href {https://doi.org/10.18653/v1/2023.emnlp-main.183} {{Enhancing
  Chat Language Models by Scaling High-quality Instructional Conversations}}.
\newblock In \emph{Proceedings of the 2023 Conference on Empirical Methods in
  Natural Language Processing}, pages 3029--3051. Association for Computational
  Linguistics.

\bibitem[{Ganesan et~al.(2025)Ganesan, Segal, Aggarwal, Sinnadurai, Lie, and
  Thangarasa}]{ganesan2025massv}
Mugilan Ganesan, Shane Segal, Ankur Aggarwal, Nish Sinnadurai, Sean Lie, and
  Vithursan Thangarasa. 2025.
\newblock \href {https://doi.org/10.18653/v1/2025.findings-emnlp.656} {{MASSV:
  Multimodal Adaptation and Self-Data Distillation for Speculative Decoding of
  Vision-Language Models}}.
\newblock In \emph{Findings of the Association for Computational Linguistics:
  EMNLP 2025}, pages 12265--12276. Association for Computational Linguistics.

\bibitem[{Goel et~al.(2024)Goel, Gagrani, Jeon, Park, Lee, and
  Lott}]{goel2024direct}
Raghavv Goel, Mukul Gagrani, Wonseok Jeon, Junyoung Park, Mingu Lee, and
  Christopher Lott. 2024.
\newblock \href {https://doi.org/10.48550/arxiv.2403.00858} {{Direct Alignment
  of Draft Model for Speculative Decoding with Chat-Fine-Tuned LLMs}}.
\newblock \emph{arXiv preprint arXiv:2403.00858}.

\bibitem[{Hinton et~al.(2015)Hinton, Vinyals, and Dean}]{hinton2015distilling}
Geoffrey Hinton, Oriol Vinyals, and Jeff Dean. 2015.
\newblock \href {https://doi.org/10.48550/arxiv.1503.02531} {{Distilling the
  Knowledge in a Neural Network}}.
\newblock \emph{arXiv preprint arXiv:1503.02531}.

\bibitem[{Hong et~al.(2025)Hong, Raju, Li, Li, Thakker, Ravichandran, Jain, and
  Hu}]{hong2025training}
Fenglu Hong, Ravi~Shanker Raju, Jonathan~Lingjie Li, Bo~Li, Urmish Thakker,
  Avinash Ravichandran, Swayambhoo Jain, and Changran Hu. 2025.
\newblock \href {https://openreview.net/forum?id=7FBJIxYXxd} {{Training Domain
  Draft Models for Speculative Decoding: Best Practices and Insights}}.
\newblock In \emph{First Workshop on Scalable Optimization for Efficient and
  Adaptive Foundation Models}.

\bibitem[{Hu et~al.(2025)Hu, Li, Xie, Lu, Toh, and Zhou}]{hu2025griffin}
Shijing Hu, Jingyang Li, Xingyu Xie, Zhihui Lu, Kim-chuan Toh, and Pan Zhou.
  2025.
\newblock \href {https://doi.org/10.52202/085713-4188} {{GRIFFIN: Effective
  Token Alignment for Faster Speculative Decoding}}.
\newblock In \emph{Advances in Neural Information Processing Systems 38}.

\bibitem[{Jain et~al.(2025)Jain, Han, Gu, Li, Yan, Zhang, Wang, Solar-Lezama,
  Sen, and Stoica}]{jain2024livecodebench}
Naman Jain, King Han, Alex Gu, Wen-Ding Li, Fanjia Yan, Tianjun Zhang, Sida
  Wang, Armando Solar-Lezama, Koushik Sen, and Ion Stoica. 2025.
\newblock \href {https://openreview.net/forum?id=chfJJYC3iL} {{LiveCodeBench:
  Holistic and Contamination Free Evaluation of Large Language Models for
  Code}}.
\newblock In \emph{International Conference on Learning Representations}.

\bibitem[{Kang et~al.(2026)Kang, Lee, Seo, and Lim}]{kang2026thinkfuse}
Myunghoon Kang, Jungseob Lee, Jaehyung Seo, and Heuiseok Lim. 2026.
\newblock \href {https://arxiv.org/abs/2610.07803} {{ThinkFuse:
  Trajectory-Aware Test-Time Fusion for Small Reasoning Models}}.
\newblock \emph{arXiv preprint arXiv:2610.07803}.

\bibitem[{Kim and Rush(2016)}]{kim2016sequence}
Yoon Kim and Alexander~M. Rush. 2016.
\newblock \href {https://doi.org/10.18653/v1/d16-1139} {{Sequence-Level
  Knowledge Distillation}}.
\newblock In \emph{Proceedings of the 2016 Conference on Empirical Methods in
  Natural Language Processing}, pages 1317--1327. Association for Computational
  Linguistics.

\bibitem[{Kwon et~al.(2023)Kwon, Li, Zhuang, Sheng, Zheng, Yu, Gonzalez, Zhang,
  and Stoica}]{kwon2023efficient}
Woosuk Kwon, Zhuohan Li, Siyuan Zhuang, Ying Sheng, Lianmin Zheng, Cody~Hao Yu,
  Joseph Gonzalez, Hao Zhang, and Ion Stoica. 2023.
\newblock \href {https://doi.org/10.1145/3600006.3613165} {{Efficient Memory
  Management for Large Language Model Serving with PagedAttention}}.
\newblock In \emph{{Proceedings of the 29th Symposium on Operating Systems
  Principles}}.

\bibitem[{Lee and Eo(2026)}]{lee2026cast}
Jungseob Lee and Sugyeong Eo. 2026.
\newblock \href {https://arxiv.org/abs/2610.00321} {{CAST: Cost-Aware
  Speculative Trees from One-Pass Block Drafters}}.
\newblock \emph{arXiv preprint arXiv:2610.00321}.

\bibitem[{Lee et~al.(2026{\natexlab{a}})Lee, Eo, Hong, Lee, Park, Seo, and
  Lim}]{lee2026directional}
Jungseob Lee, Sugyeong Eo, Seongtae Hong, Seungyoon Lee, Chanjun Park, Jaehyung
  Seo, and Heuiseok Lim. 2026{\natexlab{a}}.
\newblock \href {https://arxiv.org/abs/2610.00997} {{Distilling Directional
  Verification}}.
\newblock \emph{arXiv preprint arXiv:2610.00997}.

\bibitem[{Lee et~al.(2026{\natexlab{b}})Lee, Hong, Lee, Park, Seo, Eo, and
  Lim}]{lee2026glance}
Jungseob Lee, Seongtae Hong, Dongyub~Jude Lee, Chanjun Park, Jaehyung Seo,
  Sugyeong Eo, and Heuiseok Lim. 2026{\natexlab{b}}.
\newblock \href {https://arxiv.org/abs/2609.00355} {{Vision Is Not Overhead:
  One-Pass Block Drafting for Lossless Speculative Decoding in Vision-Language
  Models}}.
\newblock \emph{arXiv preprint arXiv:2609.00355}.

\bibitem[{Lee et~al.(2026{\natexlab{c}})Lee, Hong, Lee, Seo, Son, Eo, Park,
  Park, Moon, and Lim}]{lee2026dart_routing}
Jungseob Lee, Seongtae Hong, Seungjun Lee, Jaehyung Seo, Junyoung Son, Sugyeong
  Eo, Chanjun Park, Hyeongju Park, Hyeonseok Moon, and Heuiseok Lim.
  2026{\natexlab{c}}.
\newblock \href {https://arxiv.org/abs/2606.23181} {{DART: Draft-Agreement
  Routing for Training-Free Adaptive Thinking Budgets in Hybrid Reasoning
  Models}}.
\newblock \emph{arXiv preprint arXiv:2606.23181}.

\bibitem[{Lee et~al.(2025)Lee, Hong, Moon, and Lim}]{lee2025clo}
Jungseob Lee, Seongtae Hong, Hyeonseok Moon, and Heuiseok Lim. 2025.
\newblock \href {https://doi.org/10.18653/v1/2025.acl-long.734} {{Cross-Lingual
  Optimization for Language Transfer in Large Language Models}}.
\newblock In \emph{Proceedings of the 63rd Annual Meeting of the Association
  for Computational Linguistics (Volume 1: Long Papers)}, pages 15100--15119.
  Association for Computational Linguistics.

\bibitem[{Lee et~al.(2026{\natexlab{d}})Lee, Lee, Park, Eo, and
  Lim}]{lee2026redline}
Jungseob Lee, Dongyub~Jude Lee, Chanjun Park, Sugyeong Eo, and Heuiseok Lim.
  2026{\natexlab{d}}.
\newblock \href {https://arxiv.org/abs/2609.33887} {{Faster Block-Diffusion
  Serving with Distribution-Free Risk Guarantees}}.
\newblock \emph{arXiv preprint arXiv:2609.33887}.

\bibitem[{Lee et~al.(2026{\natexlab{e}})Lee, Lee, Hong, Eo, and
  Lim}]{lee2026bytecross}
Jungseob Lee, Seungyoon Lee, Seongtae Hong, Sugyeong Eo, and Heuiseok Lim.
  2026{\natexlab{e}}.
\newblock \href {https://arxiv.org/abs/2609.33889} {{Where Activation Sparsity
  and KV-Cache Sparsity Cross in LLM Decoding}}.
\newblock \emph{arXiv preprint arXiv:2609.33889}.

\bibitem[{Lee et~al.(2026{\natexlab{f}})Lee, Lee, Hong, Kim, Park, and
  Lim}]{lee2026acoer}
Jungseob Lee, Seungyoon Lee, Seongtae Hong, Minhyuk Kim, Chanjun Park, and
  Heuiseok Lim. 2026{\natexlab{f}}.
\newblock \href {https://arxiv.org/abs/2606.22716} {{Beyond Penalizing
  Mistakes: Stabilizing Efficiency Training in Large Reasoning Models via
  Adaptive Correct-Only Rewards}}.
\newblock \emph{arXiv preprint arXiv:2606.22716}.

\bibitem[{Lee et~al.(2026{\natexlab{g}})Lee, Lee, Son, Lee, Han, Eo, and
  Lim}]{lee2026answerconditioned}
Jungseob Lee, Seungyoon Lee, Suhyune Son, Dongyub~Jude Lee, Sungbin Han,
  Sugyeong Eo, and Heuiseok Lim. 2026{\natexlab{g}}.
\newblock \href {https://doi.org/10.48550/arxiv.2607.14552}
  {{Answer-Conditioned Chains of Thought Degrade Verifiable-Reasoning
  Distillation in Large Language Models}}.
\newblock \emph{arXiv preprint arXiv:2607.14552}.

\bibitem[{Lee et~al.(2026{\natexlab{h}})Lee, Park, Eo, and
  Moon}]{lee2026recovering}
Jungseob Lee, Chanjun Park, Sugyeong Eo, and Hyeonseok Moon.
  2026{\natexlab{h}}.
\newblock \href {https://arxiv.org/abs/2609.38795} {{Recovering Off-Policy
  Supervision for Speculative Decoding}}.
\newblock \emph{arXiv preprint arXiv:2609.38795}.

\bibitem[{Leviathan et~al.(2023)Leviathan, Kalman, and
  Matias}]{leviathan2023fast}
Yaniv Leviathan, Matan Kalman, and Yossi Matias. 2023.
\newblock \href {https://proceedings.mlr.press/v202/leviathan23a.html} {{Fast
  Inference from Transformers via Speculative Decoding}}.
\newblock In \emph{Proceedings of the 40th International Conference on Machine
  Learning}, volume 202 of \emph{Proceedings of Machine Learning Research},
  pages 19274--19286. PMLR.

\bibitem[{Li et~al.(2026)Li, Wang, Zhu, Wang, Yin, Shi, Chen, Dong, Chen, Pan,
  Li, Xie, Zhang, Yu, Wen, Tsang, and Zhang}]{li2026specforge}
Shenggui Li, Chao Wang, Yikai Zhu, Yubo Wang, Fan Yin, Shuai Shi, Yefei Chen,
  Xiaomin Dong, Qiaoling Chen, Jin Pan, Ji~Li, Laixin Xie, Yineng Zhang, Lei
  Yu, Yonggang Wen, Ivor Tsang, and Tianwei Zhang. 2026.
\newblock \href {https://arxiv.org/abs/2603.18567} {{SpecForge: A Flexible and
  Efficient Open-Source Training Framework for Speculative Decoding}}.
\newblock \emph{arXiv preprint arXiv:2603.18567}.

\bibitem[{Li et~al.(2024{\natexlab{a}})Li, Wei, Zhang, and
  Zhang}]{li2024eagle2}
Yuhui Li, Fangyun Wei, Chao Zhang, and Hongyang Zhang. 2024{\natexlab{a}}.
\newblock \href {https://doi.org/10.18653/v1/2024.emnlp-main.422} {{EAGLE-2:
  Faster Inference of Language Models with Dynamic Draft Trees}}.
\newblock In \emph{{Proceedings of the 2024 Conference on Empirical Methods in
  Natural Language Processing}}.

\bibitem[{Li et~al.(2024{\natexlab{b}})Li, Wei, Zhang, and Zhang}]{li2024eagle}
Yuhui Li, Fangyun Wei, Chao Zhang, and Hongyang Zhang. 2024{\natexlab{b}}.
\newblock \href {https://proceedings.mlr.press/v235/li24bt.html} {{EAGLE:
  Speculative Sampling Requires Rethinking Feature Uncertainty}}.
\newblock In \emph{International Conference on Machine Learning}.

\bibitem[{Li et~al.(2025)Li, Wei, Zhang, and Zhang}]{li2025eagle3}
Yuhui Li, Fangyun Wei, Chao Zhang, and Hongyang Zhang. 2025.
\newblock \href {https://doi.org/10.52202/085713-4562} {{EAGLE-3: Scaling up
  Inference Acceleration of Large Language Models via Training-Time Test}}.
\newblock In \emph{Advances in Neural Information Processing Systems 38}.

\bibitem[{Lin et~al.(2014)Lin, Maire, Belongie, Hays, Perona, Ramanan, Dollár,
  and Zitnick}]{lin2014coco}
Tsung-Yi Lin, Michael Maire, Serge Belongie, James Hays, Pietro Perona, Deva
  Ramanan, Piotr Dollár, and C.~Lawrence Zitnick. 2014.
\newblock \href {https://doi.org/10.1007/978-3-319-10602-1_48} {{Microsoft
  COCO: Common Objects in Context}}.
\newblock In \emph{Computer Vision -- ECCV 2014}, volume 8693 of \emph{Lecture
  Notes in Computer Science}, pages 740--755. Springer.

\bibitem[{Liu et~al.(2024)Liu, Hu, Bailis, Cheung, Deng, Stoica, and
  Zhang}]{liu2023online}
Xiaoxuan Liu, Lanxiang Hu, Peter Bailis, Alvin Cheung, Zhijie Deng, Ion Stoica,
  and Hao Zhang. 2024.
\newblock \href {https://proceedings.mlr.press/v235/liu24y.html} {{Online
  Speculative Decoding}}.
\newblock In \emph{International Conference on Machine Learning}.

\bibitem[{Mathew et~al.(2021)Mathew, Karatzas, and Jawahar}]{mathew2021docvqa}
Minesh Mathew, Dimosthenis Karatzas, and C.~V. Jawahar. 2021.
\newblock \href {https://doi.org/10.1109/wacv48630.2021.00225} {{DocVQA: A
  Dataset for VQA on Document Images}}.
\newblock In \emph{2021 IEEE Winter Conference on Applications of Computer
  Vision (WACV)}, pages 2199--2208. IEEE.

\bibitem[{Singh et~al.(2019)Singh, Natarajan, Shah, Jiang, Chen, Batra, Parikh,
  and Rohrbach}]{singh2019textvqa}
Amanpreet Singh, Vivek Natarajan, Meet Shah, Yu~Jiang, Xinlei Chen, Dhruv
  Batra, Devi Parikh, and Marcus Rohrbach. 2019.
\newblock \href {https://doi.org/10.1109/cvpr.2019.00851} {{Towards VQA Models
  That Can Read}}.
\newblock In \emph{2019 IEEE/CVF Conference on Computer Vision and Pattern
  Recognition (CVPR)}, pages 8309--8318. IEEE.

\bibitem[{Taori et~al.(2023)Taori, Gulrajani, Zhang, Dubois, Li, Guestrin,
  Liang, and Hashimoto}]{taori2023alpaca}
Rohan Taori, Ishaan Gulrajani, Tianyi Zhang, Yann Dubois, Xuechen Li, Carlos
  Guestrin, Percy Liang, and Tatsunori~B. Hashimoto. 2023.
\newblock \href {https://github.com/tatsu-lab/stanford_alpaca} {{Stanford
  Alpaca: An Instruction-following LLaMA model}}.
\newblock GitHub repository.

\bibitem[{Yang et~al.(2025)Yang, Li, Yang, Zhang, Hui, Zheng, Yu, Gao, Huang,
  Lv, Zheng, Liu, Zhou, Huang, Hu, Ge, Wei, Lin, Tang, Yang, Tu, Zhang, Yang,
  Yang, Zhou, Zhou, Lin, Dang, Bao, Yang, Yu, Deng, Li, Xue, Li, Zhang, Wang,
  Zhu, Men, Gao, Liu, Luo, Li, Tang, Yin, Ren, Wang, Zhang, Ren, Fan, Su,
  Zhang, Zhang, Wan, Liu, Wang, Cui, Zhang, Zhou, and Qiu}]{yang2025qwen3}
An~Yang, Anfeng Li, Baosong Yang, Beichen Zhang, Binyuan Hui, Bo~Zheng, Bowen
  Yu, Chang Gao, Chengen Huang, Chenxu Lv, Chujie Zheng, Dayiheng Liu, Fan
  Zhou, Fei Huang, Feng Hu, Hao Ge, Haoran Wei, Huan Lin, Jialong Tang, and 41
  others. 2025.
\newblock \href {https://arxiv.org/abs/2505.09388} {{Qwen3 Technical Report}}.
\newblock \emph{arXiv preprint arXiv:2505.09388}.

\bibitem[{Zafrir et~al.(2025)Zafrir, Margulis, Shteyman, Guskin, and
  Boudoukh}]{zafrir2025fastdraft}
Ofir Zafrir, Igor Margulis, Dorin Shteyman, Shira Guskin, and Guy Boudoukh.
  2025.
\newblock \href {https://doi.org/10.18653/v1/2025.findings-acl.1156}
  {{FastDraft: How to Train Your Draft}}.
\newblock In \emph{Findings of the Association for Computational Linguistics:
  ACL 2025}, pages 22488--22505. Association for Computational Linguistics.

\bibitem[{Zhang et~al.(2025)Zhang, Wang, Huang, and Xu}]{zhang2025hass}
Lefan Zhang, Xiaodan Wang, Yanhua Huang, and Ruiwen Xu. 2025.
\newblock \href
  {https://proceedings.iclr.cc/paper_files/paper/2025/hash/575286a73f238b6516ce0467d67eadb2-Abstract-Conference.html}
  {{Learning Harmonized Representations for Speculative Sampling}}.
\newblock In \emph{International Conference on Learning Representations}.

\bibitem[{Zheng et~al.(2023)Zheng, Chiang, Sheng, Zhuang, Wu, Zhuang, Lin, Li,
  Li, Xing, Zhang, Gonzalez, and Stoica}]{zheng2023judging}
Lianmin Zheng, Wei-Lin Chiang, Ying Sheng, Siyuan Zhuang, Zhanghao Wu, Yonghao
  Zhuang, Zi~Lin, Zhuohan Li, Dacheng Li, Eric Xing, Hao Zhang, Joseph
  Gonzalez, and Ion Stoica. 2023.
\newblock \href {https://doi.org/10.52202/075280-2020} {{Judging LLM-as-a-Judge
  with MT-Bench and Chatbot Arena}}.
\newblock In \emph{Advances in Neural Information Processing Systems},
  volume~36.

\bibitem[{Zhou et~al.(2020)Zhou, Gu, and Neubig}]{zhou2019understanding}
Chunting Zhou, Jiatao Gu, and Graham Neubig. 2020.
\newblock \href {https://openreview.net/forum?id=BygFVAEKDH} {{Understanding
  Knowledge Distillation in Non-autoregressive Machine Translation}}.
\newblock In \emph{International Conference on Learning Representations}.

\bibitem[{Zhou et~al.(2024)Zhou, Lyu, Rawat, Menon, Rostamizadeh, Kumar, Kagy,
  and Agarwal}]{zhou2024distillspec}
Yongchao Zhou, Kaifeng Lyu, Ankit~Singh Rawat, Aditya~Krishna Menon, Afshin
  Rostamizadeh, Sanjiv Kumar, Jean-François Kagy, and Rishabh Agarwal. 2024.
\newblock \href
  {https://proceedings.iclr.cc/paper_files/paper/2024/hash/8766fbc68e1ed1cdef712ce273e0a363-Abstract-Conference.html}
  {{DistillSpec: Improving Speculative Decoding via Knowledge Distillation}}.
\newblock In \emph{International Conference on Learning Representations}.

\end{thebibliography}

\appendix
\raggedbottom

\section{Protocol Details}
\label{app:protocol}

\subsection{Vision training}
\label{app:vision-training}

Examples come from the ALLaVA LAION instruction file in a fixed random order. We keep the first 26,196 rows whose conversation opens with a user turn and an assistant turn of at least 40 characters and whose image belongs to the first released image archive, and we train on that first exchange. The ShareGPT4V subset takes the 30,000 GPT-4V captions of its 100K caption file whose images come from LAION, CC and SBU, sorts them by identifier, shuffles them with a fixed seed and keeps the first 26,196. On ShareGPT4V, the three objectives are trained on external and fully regenerated responses, both of which also enter the horizon study, and KD and KD + gate with $\gamma=2$ are trained once on the external responses, as Appendix~\ref{app:settings} lists. Examples with fewer than 32 supervised assistant tokens are removed. On ALLaVA this leaves 26,101 external, truncated and loss-window examples and 26,193 fully regenerated examples, so three epochs take 3,261 or 3,273 updates. No ShareGPT4V example falls below the threshold, so both ShareGPT4V versions keep all 26,196 examples and three epochs take 3,273 updates.

The drafter has five layers and reads the hidden states of target layers 1, 9, 17, 25 and 33. Training uses four GPUs with an effective batch of 24, AdamW with 4\% warmup and cosine decay, bf16, SDPA attention, gradient clipping at norm 1 and a maximum length of 4,096 tokens, and images keep between 50,176 and 200,704 pixels. The learning rate, $1.47\times10^{-3}$, is the SpecForge \citep{li2026specforge} default for DFlash training, $6\times10^{-4}$, scaled by the square root of our six times larger batch on each GPU. The KD loss uses the target distribution at temperature 1 and carries a fixed scale of 1.013, the ratio of the median gradient norms of CE and KD on identical batches before training.

Regeneration uses the target in vLLM \citep{kwon2023efficient} with greedy decoding and the image bounds and chat template of training, and the 3.8\% of regenerated ALLaVA responses that reach the 768-token limit are closed there. Regenerating the ALLaVA corpus took about 5 minutes on four H100 GPUs, on which a three-epoch ALLaVA run trains for 29 to 46 minutes. Truncation cuts the regenerated content to the token count of the external response and closes the assistant turn. The loss window keeps the full regenerated input and masks the loss after that count. In the 60 retained examples whose regenerated response is shorter than the external one, both treatments keep the whole response.

On ALLaVA, regenerated responses average 395 tokens against 112 for external ones, and a proxy for the anchors available in each example, the mean of $\min(128, L)$ over response lengths $L$, is 106.7 for external and 127.3 for fully regenerated responses. Fully regenerated ShareGPT4V responses keep the first 768 tokens of the target's greedy response, and the 3.9\% that are longer are closed there. They average 380 tokens against 203 for the external captions, and the anchor proxy is 127.5 against 126.7.

\subsection{Vision evaluation}
\label{app:vision-evaluation}

The captioning prompts are rows 100 to 199 of COCO val2017, the TextVQA prompts rows 0 to 99 of the public training split and the DocVQA prompts rows 400 to 499 of the validation split, each in release order. The first prompt of each domain is a warm-up that is not scored. No evaluation image occurs byte for byte in the training corpus. Images are resized to at most 896 pixels on the longer side.

In each round the drafter predicts the 15 slots in one forward pass, and the verifier builds a new tree from these predictions. The candidates at a slot are the drafter's eight most probable tokens there, a path is scored by the sum of the drafter's log-probabilities of its tokens, and the tree keeps the 63 best nodes, each added only after its parent. The target scores the whole tree in one forward pass with a tree attention mask and accepts the path that follows its own greedy tokens as far as the tree allows. All vision MATs come from one GPU model, the RTX A6000, in bf16.

Within each bootstrap resample, every run's MAT is recomputed from the drawn prompts, and the runs of a cell are averaged before increments and ratios are formed. MAT counts the accepted path and the target's own token in each round before the final round is clipped at the end-of-sequence token or the length limit, so it can differ from delivered tokens in that last round.

\subsection{Vision controls}
\label{app:vision-controls}

Every control uses external and fully regenerated ALLaVA responses. The controls with random initialization, one third of the learning rate and the L2 anchor train CE and KD, and the random-initialization control trains them for three and for six epochs. The control with $\gamma=2$ trains CE, KD, KD + gate and KD with the soft weighting, the control with two further data orders retrains the six factorial cells of the two versions once in each order, and the 64-token control trains KD. Apart from the data orders, control cells are trained twice. The chain verifier re-evaluates the 24 checkpoints of the external, fully regenerated and matched-length versions. The sampled evaluation re-evaluates the 12 checkpoints of the external and fully regenerated versions with the tree verifier while the target samples at temperature 1, with random numbers fixed for each prompt and output position and shared by all drafters.

The random-initialization control keeps the drafter's architecture and starts its five layers and feature projection from random weights, while the embeddings and output head come from the target as in every run. Each six-epoch run starts from the random weights and data order of its three-epoch counterpart and changes only the number of epochs, so its warmup and cosine decay span 6,522 updates on external and 6,546 on fully regenerated responses. The two further data orders shuffle the training examples into two other fixed random orders, and each of their six cells is trained once with the anchor sampling of the first factorial run, which represents the main order. The control with $\gamma=2$ keeps the rest of the recipe, and its soft weighting multiplies the target's probabilities of the preceding corpus tokens, as in PARD-2. The chain verifier runs the tree verifier's routine with one candidate at each slot and seven slots in each round, on the same prompts and GPU model. After every optimizer step, the L2 anchor moves the drafter's weights toward those of the released drafter by the current learning rate times their difference, in both objectives. The control with 64 distilled tokens keeps the rest of the KD recipe and reuses the factorial CE cells.

To separate turn closure from the other differences at the cut, we also train KD and KD + gate twice on the truncated examples with the loss masked on the tokens that close the assistant turn after the cut, keeping the examples, their order and the training length. The loss on these tokens stays in the examples that the cut leaves whole, whose closure is the target's own, and in 15 cut examples that masking would leave with fewer than 32 supervised tokens.

\subsection{Text study}
\label{app:text-protocol}

The text corpus starts from the 26,196 conversations whose messages have the lowest SHA-256 hashes in the first of the three supervised fine-tuning files of UltraChat 200k. Their first turns are kept when both responses leave at least 32 supervised tokens under the 1,024-token limit. Regenerated responses are the target's greedy responses of at most 768 new tokens. They are 1.7 times as long as the original ones, 594 against 344 supervised tokens on average, and 56.6\% reach the 768-token limit. Regenerating the text corpus took 2.9 GPU-hours on one RTX A6000, against 7.2 to 8.9 hours for each three-epoch DFlash text run on the same GPU model.

Each of the eight DFlash cells, CE, KD, KD + gate and the label control of Section~\ref{sec:analysis} on each version, is trained twice for three epochs on one RTX A6000 with the learning rate, effective batch and $\gamma$ of the vision study, using micro-batches of 3, eight accumulation steps, 3,069 updates, a KD scale of 1 and a maximum length of 1,024 tokens. Evaluation covers MT-Bench \citep{zheng2023judging}, Alpaca \citep{taori2023alpaca}, GSM8K \citep{cobbe2021training}, AIME 2024 and 2025, HumanEval \citep{chen2021evaluating} and LiveCodeBench \citep{jain2024livecodebench}. The seven tasks contribute 80, 80, 1,319, 30, 30, 164 and 400 prompts, with up to 256 new tokens. At $T=1$ the target samples and the drafter proposes its greedy tokens, and each prompt uses a fixed random seed. Tasks take the place of domains in the aggregation and the bootstrap.

\paragraph{EAGLE-3 drafter.} The second text drafter starts from a publicly released EAGLE-3 drafter for Qwen3-4B, one decoder layer that reads the fused hidden states of target layers 1, 17 and 32 and predicts over a fixed 32,000-token subset of the target vocabulary, whose mapping we keep. Each of its four cells is trained twice for three epochs with the micro-batches, accumulation steps, updates, warmup, gradient clipping and maximum length of the DFlash text runs, with seven simulated test-time steps and a learning rate of $2.45\times10^{-4}$, the SpecForge default for EAGLE-3 training, $10^{-4}$, scaled by the same factor. This drafter samples no anchors, so the two runs of a cell use two data orders. CE and KD train the positions whose corpus token lies in the drafter's vocabulary, and KD keeps the target's eight most probable tokens within that vocabulary and merges the rest, including all probability outside the vocabulary, into a ninth outcome. The fine-tuned checkpoints and the released drafter are evaluated at temperature 0 on the same 2,103 prompts with the same token limit, using the public EAGLE-3 decoder, which verifies a tree of 59 drafted tokens, at most eight deep, grown from the drafter's ten most probable tokens at each step, so its MATs are compared only within this family.

\subsection{Artifact licenses}
\label{app:licenses}

Qwen3-VL-8B-Instruct and Qwen3-4B carry Apache 2.0 and the DFlash drafters MIT. ALLaVA, ShareGPT4V and the Alpaca prompts carry CC BY-NC 4.0, UltraChat 200k, GSM8K and HumanEval carry MIT, TextVQA carries CC BY 4.0, and the MT-Bench prompts and the formatted DocVQA release carry Apache 2.0. COCO images are used under the COCO terms of use, and LiveCodeBench and the AIME problems are public benchmark releases. We use every artifact only for non-commercial research.

\section{Vision Results by Domain and Schedule}
\label{app:vision}

Table~\ref{tab:app-domains} lists MAT for every cell and domain, together with the label control of Section~\ref{sec:analysis}, and Table~\ref{tab:app-increments} lists the increments of the factorial and of ShareGPT4V for each domain. The label control adds 9.3\% [8.6, 10.0] over CE on external responses and changes MAT by $-$0.2\% [$-$0.8, 0.5] on fully regenerated ones, a ratio of ratios of $-$8.6\% [$-$9.5, $-$7.8]. KD adds 4.7\% [4.1, 5.4] over it on external and 3.9\% [3.2, 4.6] on fully regenerated responses. Between the two length controls, the ratio of ratios is 0.7\% [$-$0.3, 1.7] for KD, which is not resolved. With the turn-closing tokens masked, MAT is 3.779 under KD and 3.803 under KD + gate, a gate gain of 0.6\%. The ratio of ratios of the gate's gains is $-$1.6\% [$-$2.3, $-$0.8] against truncation and 0.4\% [$-$0.4, 1.3] against the loss window.

\begin{table}[!htb]
\centering
\fontsize{7.5}{8.5}\selectfont
\renewcommand{\arraystretch}{0.92}
\setlength{\tabcolsep}{2pt}
\begin{tabular*}{\columnwidth}{@{\extracolsep{\fill}}lcccc@{}}
\toprule
Objective & Caption & TextVQA & DocVQA & Mean \\
\midrule
Initialization & 2.495 & 2.806 & 3.254 & 2.835 \\
\rowcolor{groupyellow}[0pt][0pt]\multicolumn{5}{@{}c@{}}{\textit{ALLaVA, external}\strut} \\
CE & 2.565 & 2.650 & 3.498 & 2.875 \\
KD & 2.915 & 3.136 & 3.898 & 3.291 \\
KD + gate & 3.067 & 3.313 & 4.052 & 3.453 \\
Greedy labels & 2.762 & 3.008 & 3.736 & 3.143 \\
\rowcolor{groupyellow}[0pt][0pt]\multicolumn{5}{@{}c@{}}{\textit{ALLaVA, full regeneration}\strut} \\
CE & 3.377 & 3.940 & 4.267 & 3.843 \\
KD & 3.506 & 4.097 & 4.411 & 3.987 \\
KD + gate & 3.500 & 4.085 & 4.360 & 3.965 \\
Greedy labels & 3.382 & 3.951 & 4.227 & 3.837 \\
\rowcolor{groupyellow}[0pt][0pt]\multicolumn{5}{@{}c@{}}{\textit{ALLaVA, truncated}\strut} \\
CE & 3.141 & 3.457 & 3.890 & 3.483 \\
KD & 3.312 & 3.775 & 4.131 & 3.724 \\
KD + gate & 3.348 & 3.881 & 4.246 & 3.807 \\
\rowcolor{groupyellow}[0pt][0pt]\multicolumn{5}{@{}c@{}}{\textit{ALLaVA, loss window}\strut} \\
CE & 3.154 & 3.636 & 3.960 & 3.568 \\
KD & 3.344 & 3.880 & 4.190 & 3.789 \\
KD + gate & 3.344 & 3.883 & 4.216 & 3.797 \\
\rowcolor{groupyellow}[0pt][0pt]\multicolumn{5}{@{}c@{}}{\textit{ShareGPT4V, external}\strut} \\
CE & 2.542 & 2.691 & 3.216 & 2.802 \\
KD & 3.112 & 3.110 & 3.735 & 3.307 \\
KD + gate & 3.357 & 3.269 & 3.930 & 3.507 \\
\rowcolor{groupyellow}[0pt][0pt]\multicolumn{5}{@{}c@{}}{\textit{ShareGPT4V, full regeneration}\strut} \\
CE & 4.097 & 3.630 & 3.904 & 3.872 \\
KD & 4.205 & 3.775 & 4.072 & 4.013 \\
KD + gate & 4.205 & 3.817 & 4.141 & 4.051 \\
\bottomrule
\end{tabular*}

\caption{Vision MAT for each domain, averaged over two training runs. The last column is the geometric mean over domains. Greedy labels trains CE toward the target's greedy token.}
\label{tab:app-domains}
\end{table}

When the two training runs of each cell are resampled along with the prompts, the intervals of the full-regeneration ratios of ratios widen to [$-$10.6, $-$8.1] for KD and [$-$6.2, $-$4.2] for the gate. On ShareGPT4V they are [$-$13.3, $-$11.0] for KD and [$-$6.0, $-$3.7] for the gate. With the runs resampled, every ratio of ratios in Figure~\ref{fig:theta} stays below zero, the highest upper bound being $-$1.3\%. The gate's ratio of ratios between truncation and the loss window becomes 2.0\% [0.7, 3.4], the gate's increment on responses sampled at temperature 1 becomes 2.1\% [0.9, 3.3], and the KD contrast between the length controls becomes 0.7\% [$-$0.6, 2.0], which stays unresolved.

Table~\ref{tab:horizon} lists the CE and KD schedules of the horizon study, Table~\ref{tab:app-horizon} gives their domain values, and Table~\ref{tab:app-horizon-gate} lists the gate schedules. The bootstrap holds each best schedule fixed. Relative to half an epoch, a quarter epoch changes external ALLaVA CE by $-$1.2\% [$-$2.1, $-$0.4]. On ShareGPT4V the best measured schedules are half an epoch for CE and KD on external responses and one epoch for CE and three for KD on regenerated ones. When they are instead selected on alternate prompts of each domain and the ratio of ratios is estimated on the remaining prompts, the two halves select different regenerated KD schedules, one and three epochs, whose MATs differ by 0.03\%, and the best-schedule ratio of ratios averaged over the two directions is $-$8.6\%. For the gate over KD on the schedules of Table~\ref{tab:app-horizon-gate}, both halves select the same schedules and the averaged ratio of ratios is $-$3.2\%.

\begin{table}[!htb]
\centering
\fontsize{7.5}{8.5}\selectfont
\renewcommand{\arraystretch}{0.92}
\setlength{\tabcolsep}{2pt}
\begin{tabular*}{\columnwidth}{@{\extracolsep{\fill}}lccc@{}}
\toprule
Epochs & Caption & TextVQA & DocVQA \\
\midrule
\rowcolor{groupyellow}[0pt][0pt]\multicolumn{4}{@{}c@{}}{\textit{ALLaVA, external responses}\strut} \\
0.25 & 2.718 / 2.879 & 2.887 / 3.194 & 3.789 / 3.971 \\
0.5 & 2.747 / 2.922 & 2.928 / 3.247 & 3.837 / 3.990 \\
1 & 2.761 / 2.980 & 2.894 / 3.240 & 3.846 / 4.071 \\
2 & 2.723 / 2.983 & 2.849 / 3.224 & 3.758 / 4.059 \\
3 & 2.565 / 2.915 & 2.650 / 3.136 & 3.498 / 3.898 \\
\rowcolor{groupyellow}[0pt][0pt]\multicolumn{4}{@{}c@{}}{\textit{ShareGPT4V, external responses}\strut} \\
0.25 & 2.730 / 3.053 & 2.982 / 3.243 & 3.804 / 4.014 \\
0.5 & 2.747 / 3.119 & 3.007 / 3.282 & 3.764 / 4.041 \\
1 & 2.727 / 3.151 & 2.968 / 3.221 & 3.694 / 3.991 \\
2 & 2.673 / 3.162 & 2.877 / 3.202 & 3.520 / 3.890 \\
3 & 2.542 / 3.112 & 2.691 / 3.110 & 3.216 / 3.735 \\
\rowcolor{groupyellow}[0pt][0pt]\multicolumn{4}{@{}c@{}}{\textit{ALLaVA, target regeneration}\strut} \\
1 & 3.362 / 3.390 & 3.922 / 3.947 & 4.360 / 4.412 \\
2 & 3.417 / 3.473 & 3.993 / 4.102 & 4.331 / 4.390 \\
3 & 3.377 / 3.506 & 3.940 / 4.097 & 4.267 / 4.411 \\
4 & 3.303 / 3.452 & 3.811 / 4.072 & 4.083 / 4.308 \\
6 & 3.091 / 3.383 & 3.583 / 3.956 & 3.844 / 4.188 \\
\rowcolor{groupyellow}[0pt][0pt]\multicolumn{4}{@{}c@{}}{\textit{ShareGPT4V, target regeneration}\strut} \\
0.5 & 3.787 / 3.811 & 3.724 / 3.688 & 4.190 / 4.194 \\
1 & 3.961 / 3.992 & 3.802 / 3.828 & 4.199 / 4.227 \\
3 & 4.097 / 4.205 & 3.630 / 3.775 & 3.904 / 4.072 \\
6 & 3.844 / 4.148 & 3.215 / 3.601 & 3.467 / 3.783 \\
\bottomrule
\end{tabular*}

\caption{Vision MAT for each domain, corpus, response source and training length, averaged over two training runs. Each cell gives CE / KD.}
\label{tab:app-horizon}
\end{table}

\begin{table}[!htb]
\centering
\fontsize{7.5}{8.5}\selectfont
\renewcommand{\arraystretch}{0.92}
\setlength{\tabcolsep}{2pt}
\begin{tabular*}{\columnwidth}{@{\extracolsep{\fill}}lccc@{}}
\toprule
Epochs & KD & KD + gate & Gate gain (\%) \\
\midrule
\rowcolor{groupyellow}[0pt][0pt]\multicolumn{4}{@{}c@{}}{\textit{ALLaVA, external responses}\strut} \\
0.5 & 3.357 & 3.454 & +2.9 \\
1 & 3.400 & 3.492 & +2.7 \\
2 & 3.392 & 3.488 & +2.8 \\
3 & 3.291 & 3.453 & +4.9 \\
\rowcolor{groupyellow}[0pt][0pt]\multicolumn{4}{@{}c@{}}{\textit{ALLaVA, target regeneration}\strut} \\
1 & 3.894 & 3.881 & $-$0.3 \\
3 & 3.987 & 3.965 & $-$0.5 \\
6 & 3.827 & 3.882 & +1.5 \\
\bottomrule
\end{tabular*}

\caption{Vision MAT of KD and of KD with the survival gate on ALLaVA for each response source and training length, with the gate's gain over KD. Every row is a separately trained and annealed schedule.}
\label{tab:app-horizon-gate}
\end{table}

\begin{table}[!htb]
\centering
\fontsize{7.5}{8.5}\selectfont
\renewcommand{\arraystretch}{0.92}
\setlength{\tabcolsep}{1.5pt}
\begin{tabular*}{\columnwidth}{@{\extracolsep{\fill}}lcccc@{}}
\toprule
Gain & Mean [95\% CI] & Caption & TextVQA & DocVQA \\
\midrule
\rowcolor{groupyellow}[0pt][0pt]\multicolumn{5}{@{}c@{}}{\textit{ALLaVA, external}\strut} \\
KD & +14.4 [13.7, 15.2] & +13.7 & +18.3 & +11.4 \\
Gate & +4.9 [4.3, 5.6] & +5.2 & +5.6 & +3.9 \\
\rowcolor{groupyellow}[0pt][0pt]\multicolumn{5}{@{}c@{}}{\textit{ALLaVA, full regeneration}\strut} \\
KD & +3.7 [3.0, 4.5] & +3.8 & +4.0 & +3.4 \\
Gate & $-$0.5 [$-$1.1, 0.0] & $-$0.2 & $-$0.3 & $-$1.1 \\
\rowcolor{groupyellow}[0pt][0pt]\multicolumn{5}{@{}c@{}}{\textit{ALLaVA, truncated}\strut} \\
KD & +6.9 [6.2, 7.7] & +5.4 & +9.2 & +6.2 \\
Gate & +2.2 [1.6, 2.9] & +1.1 & +2.8 & +2.8 \\
\rowcolor{groupyellow}[0pt][0pt]\multicolumn{5}{@{}c@{}}{\textit{ALLaVA, loss window}\strut} \\
KD & +6.2 [5.4, 6.9] & +6.0 & +6.7 & +5.8 \\
Gate & +0.2 [$-$0.4, 0.8] & 0.0 & +0.1 & +0.6 \\
\rowcolor{groupyellow}[0pt][0pt]\multicolumn{5}{@{}c@{}}{\textit{ShareGPT4V, external}\strut} \\
KD & +18.0 [17.2, 18.8] & +22.4 & +15.6 & +16.1 \\
Gate & +6.1 [5.3, 6.8] & +7.9 & +5.1 & +5.2 \\
\rowcolor{groupyellow}[0pt][0pt]\multicolumn{5}{@{}c@{}}{\textit{ShareGPT4V, full regeneration}\strut} \\
KD & +3.6 [3.0, 4.3] & +2.6 & +4.0 & +4.3 \\
Gate & +0.9 [0.3, 1.5] & 0.0 & +1.1 & +1.7 \\
\bottomrule
\end{tabular*}

\caption{Relative MAT increments (\%), overall with 95\% prompt bootstrap intervals and for each domain. KD compares with CE, and gate compares with KD.}
\label{tab:app-increments}
\end{table}

\section{Controls of the Vision Results}
\label{app:controls}

Table~\ref{tab:app-theta} lists every ratio of ratios, including those plotted in Figure~\ref{fig:theta}, and each of its data-order rows comes from one training run. In Table~\ref{tab:cekd}, every cell trained for three epochs from random initialization is 26\% to 31\% below the same cell trained from the released drafter in Table~\ref{tab:main}a, and its KD ratio of ratios is smaller in magnitude, $-$5.8\% against $-$9.4\%, both after three epochs. Trained twice as long, the control's KD ratio of ratios, $-$8.5\%, is negative in every domain and training run, and its regenerated CE exceeds the released drafter by 8.7\%.

\begin{table}[!htb]
\centering
\fontsize{7.5}{8.5}\selectfont
\renewcommand{\arraystretch}{0.92}
\setlength{\tabcolsep}{1.5pt}
\begin{tabular*}{\columnwidth}{@{\extracolsep{\fill}}llc@{}}
\toprule
Setting & Increment & Estimate [95\% CI] \\
\midrule
\rowcolor{groupyellow}[0pt][0pt]\multicolumn{3}{@{}c@{}}{\textit{Vision, tree verifier}\strut} \\
Full regeneration & KD & $-$9.4 [$-$10.2, $-$8.5] \\
Full regeneration & Gate & $-$5.2 [$-$6.0, $-$4.4] \\
Full regeneration, $T=1$ & KD & $-$8.2 [$-$9.1, $-$7.3] \\
Full regeneration, $T=1$ & Gate & $-$4.0 [$-$4.8, $-$3.1] \\
Truncated & KD & $-$6.6 [$-$7.5, $-$5.6] \\
Truncated & Gate & $-$2.6 [$-$3.4, $-$1.8] \\
Loss window & KD & $-$7.2 [$-$8.1, $-$6.3] \\
Loss window & Gate & $-$4.5 [$-$5.3, $-$3.7] \\
Sampled, $T=1$ & KD & $-$8.9 [$-$9.7, $-$8.2] \\
Sampled, $T=1$ & Gate & $-$2.7 [$-$3.5, $-$1.9] \\
Sampled, $T=0.7$ & KD & $-$9.5 [$-$10.3, $-$8.7] \\
Sampled, $T=0.7$ & Gate & $-$3.7 [$-$4.5, $-$2.8] \\
ShareGPT4V, full regeneration & KD & $-$12.2 [$-$13.0, $-$11.4] \\
ShareGPT4V, full regeneration & Gate & $-$4.8 [$-$5.7, $-$4.0] \\
\rowcolor{groupyellow}[0pt][0pt]\multicolumn{3}{@{}c@{}}{\textit{Vision, chain verifier}\strut} \\
Full regeneration & KD & $-$8.9 [$-$9.6, $-$8.1] \\
Full regeneration & Gate & $-$4.9 [$-$5.7, $-$4.2] \\
Truncated & KD & $-$6.4 [$-$7.2, $-$5.6] \\
Truncated & Gate & $-$4.6 [$-$5.3, $-$3.8] \\
Loss window & KD & $-$7.2 [$-$8.0, $-$6.4] \\
Loss window & Gate & $-$5.4 [$-$6.2, $-$4.6] \\
\rowcolor{groupyellow}[0pt][0pt]\multicolumn{3}{@{}c@{}}{\textit{Vision controls, full regeneration}\strut} \\
Random initialization, three epochs & KD & $-$5.8 [$-$6.5, $-$5.2] \\
Random initialization, six epochs & KD & $-$8.5 [$-$9.3, $-$7.7] \\
$\gamma=2$ & KD & $-$10.4 [$-$11.2, $-$9.5] \\
$\gamma=2$ & Gate & $-$4.6 [$-$5.5, $-$3.8] \\
$\gamma=2$ & Soft weight & $-$4.5 [$-$5.4, $-$3.7] \\
Second data order & KD & $-$10.0 [$-$11.0, $-$8.9] \\
Second data order & Gate & $-$3.9 [$-$5.0, $-$2.7] \\
Third data order & KD & $-$10.4 [$-$11.5, $-$9.3] \\
Third data order & Gate & $-$3.8 [$-$5.0, $-$2.5] \\
Lower learning rate & KD & $-$8.3 [$-$9.2, $-$7.5] \\
L2 anchor & KD & $-$7.7 [$-$8.5, $-$6.9] \\
64 distilled tokens & KD & $-$9.9 [$-$10.8, $-$9.1] \\
Best schedules & KD & $-$5.6 [$-$6.4, $-$4.8] \\
Best schedules & Gate & $-$3.1 [$-$4.0, $-$2.3] \\
ShareGPT4V, best schedules & KD & $-$8.4 [$-$9.4, $-$7.4] \\
\rowcolor{groupyellow}[0pt][0pt]\multicolumn{3}{@{}c@{}}{\textit{Text replication}\strut} \\
DFlash, $T=0$ & KD & $-$5.9 [$-$6.4, $-$5.4] \\
DFlash, $T=0$ & Gate & $-$4.3 [$-$4.7, $-$3.8] \\
DFlash, $T=1$ & KD & $-$5.5 [$-$6.8, $-$4.2] \\
DFlash, $T=1$ & Gate & $-$4.2 [$-$5.7, $-$2.7] \\
EAGLE-3, $T=0$ & KD & $-$10.2 [$-$10.7, $-$9.6] \\
\bottomrule
\end{tabular*}

\caption{Ratios of ratios (\%) with 95\% prompt bootstrap intervals, comparing each increment on the given responses with its counterpart on external responses. KD compares with CE. Gate and soft weight compare with KD.}
\label{tab:app-theta}
\end{table}

With $\gamma=2$, KD with the soft survival weighting of PARD-2 in place of the gate reaches 3.542 on external and 4.053 on fully regenerated responses, close to KD + gate in Table~\ref{tab:main}b. The lower learning rate, $4.9\times10^{-4}$, raises external CE and KD by 7.8\% and 5.6\% over the factorial and changes fully regenerated CE and KD by 1.3\% and 0.3\%. The L2 anchor holds every run to a relative weight drift of 0.14 to 0.16 from the released drafter, about half that of the factorial runs, and leaves 81\% [71, 93] of the KD shrinkage on the log scale. Distilling the target's 64 most probable tokens in place of eight changes the MAT of KD trained on fully regenerated responses by $-$0.3\% [$-$0.9, 0.4] and gives a KD ratio of ratios of $-$9.9\%, against $-$9.4\% with eight tokens.

Every chain-verifier MAT in Table~\ref{tab:main}b is lower than the tree-verifier MAT of the same cell in Table~\ref{tab:main}a because each round checks one candidate at each of seven slots. Table~\ref{tab:app-rd} lists the three data orders, in which the gate changes MAT after regeneration by $-$0.4\%, 0.2\% and 0.8\%, each with an interval that includes zero. Across the three orders, the ratios of ratios have standard deviations of 0.5 points for KD and 0.7 points for the gate.

\begin{table}[!htb]
\centering
\fontsize{7.5}{8.5}\selectfont
\renewcommand{\arraystretch}{0.92}
\setlength{\tabcolsep}{1.5pt}
\begin{tabular*}{\columnwidth}{@{\extracolsep{\fill}}lccc@{}}
\toprule
\multicolumn{4}{@{}c@{}}{(a) MAT} \\
Responses & CE & KD & KD + gate \\
\midrule
\rowcolor{groupyellow}[0pt][0pt]\multicolumn{4}{@{}c@{}}{\textit{Main data order}\strut} \\
External & 2.882 & 3.286 & 3.448 \\
Full regeneration & 3.864 & 3.988 & 3.973 \\
\rowcolor{groupyellow}[0pt][0pt]\multicolumn{4}{@{}c@{}}{\textit{Second data order}\strut} \\
External & 2.872 & 3.307 & 3.448 \\
Full regeneration & 3.832 & 3.972 & 3.982 \\
\rowcolor{groupyellow}[0pt][0pt]\multicolumn{4}{@{}c@{}}{\textit{Third data order}\strut} \\
External & 2.872 & 3.314 & 3.471 \\
Full regeneration & 3.835 & 3.966 & 3.996 \\
\bottomrule
\end{tabular*}
\par\smallskip
\begin{tabular*}{\columnwidth}{@{\extracolsep{\fill}}lcc@{}}
\toprule
\multicolumn{3}{@{}c@{}}{(b) Ratio of ratios (\%)} \\
Data order & KD [95\% CI] & Gate [95\% CI] \\
\midrule
Main & $-$9.5 [$-$10.5, $-$8.5] & $-$5.0 [$-$6.0, $-$4.0] \\
Second & $-$10.0 [$-$11.0, $-$8.9] & $-$3.9 [$-$5.0, $-$2.7] \\
Third & $-$10.4 [$-$11.5, $-$9.3] & $-$3.8 [$-$5.0, $-$2.5] \\
\bottomrule
\end{tabular*}

\caption{Vision MAT and ratios of ratios with 95\% intervals in three data orders, from one training run in each order. The main order is the factorial's, represented by its first run.}
\label{tab:app-rd}
\end{table}

On responses sampled at temperature 1, we also train CE and KD, twice each, on three target samples of every prompt, with the recipe otherwise unchanged. The first sample is that sampled version, and the other two are drawn in the same way: each sample draws every response with one random seed shared across prompts. Training runs for the 3,273 updates that three epochs take on one sample. This is one pass over the three samples, so every prompt is seen about three times, each time with a different response. For this comparison, KD on one sample is retrained alongside the new cells.

CE and KD reach 3.977 and 4.022 on three samples, against 3.756 and 3.926 on one. Three samples raise CE by 5.9\% [5.1, 6.6] and KD by 2.4\% [1.8, 3.1], so KD's gain over CE falls from 4.5\% to 1.1\% [0.6, 1.6], a ratio of ratios of $-$3.2\% [$-$4.0, $-$2.5]. Three samples therefore remove 0.75 [0.63, 0.86] of KD's gain. When the two training runs of each cell are resampled along with the prompts, the intervals of the two increases, the three-sample gain and the ratio of ratios become [4.4, 7.3], [1.4, 3.5], [0.2, 2.1] and [$-$4.8, $-$1.6]. The KD cell sampled at $T=1$ in Table~\ref{tab:main}a differs from the retrained one by $-$0.3\% [$-$0.9, 0.3]; with it, the one-sample gain is 4.2\% and the ratio of ratios is $-$3.0\% [$-$3.7, $-$2.2].

\section{Additional Analyses}
\label{app:analysis}

Table~\ref{tab:app-mix} lists the mixtures of Figure~\ref{fig:dose}. Every mixture keeps the 26,101 external examples and their order, and the fully regenerated point is the factorial cell, which keeps its own 26,193 examples and order. Both gains fall in all three domains at the first step, and at each later step at least one of them rises in some domain.

\begin{table}[!htb]
\centering
\fontsize{7.5}{8.5}\selectfont
\renewcommand{\arraystretch}{0.92}
\setlength{\tabcolsep}{2pt}
\begin{tabular*}{\columnwidth}{@{\extracolsep{\fill}}lcccccc@{}}
\toprule
\multicolumn{2}{@{}c}{Corpus (\%)} & \multicolumn{3}{c}{MAT} & \multicolumn{2}{c}{Gain (\%)} \\
\cmidrule(r){1-2}\cmidrule(lr){3-5}\cmidrule(l){6-7}
Share & Agreement & CE & KD & KD + gate & KD & Gate \\
\midrule
0 & 58.0 & 2.875 & 3.291 & 3.453 & +14.4 & +4.9 \\
25 & 68.9 & 3.468 & 3.724 & 3.820 & +7.4 & +2.6 \\
50 & 78.8 & 3.625 & 3.829 & 3.880 & +5.6 & +1.4 \\
75 & 87.5 & 3.714 & 3.907 & 3.932 & +5.2 & +0.6 \\
100 & 95.2 & 3.843 & 3.987 & 3.965 & +3.7 & $-$0.5 \\
\bottomrule
\end{tabular*}

\caption{Vision MAT on ALLaVA when a share of the external responses is replaced by the target's greedy regenerations, averaged over two training runs, with agreement and the gains of KD over CE and of the gate over KD.}
\label{tab:app-mix}
\end{table}

Table~\ref{tab:app-slots} lists KD's increments over CE in the top-1 agreement and the top-8 recall of the target's greedy token at each block slot, scored on the target's greedy text for the 297 MAT prompts. Their mean difference over slots 1 to 7, top-8 minus top-1, is 1.6 points [1.4, 1.8] on external responses and 0.3 points [0.2, 0.4] after full regeneration. The chain's share of KD's tree gain, the ratio of $\log(1+I_c)$ to $\log(1+I_t)$ for KD's increments over CE scored as a seven-slot chain, $I_c$, and as the tree, $I_t$, is 0.78 [0.73, 0.83] on external responses and 0.28 [0.14, 0.43] after full regeneration. A chain of the drafter's first choices thus keeps most of KD's tree gain on external responses, where KD mostly corrects the first choice, but less than half after full regeneration, where most of the remaining gain lies in how the other tree candidates are ranked.

\begin{table}[!htb]
\centering
\fontsize{7.5}{8.5}\selectfont
\renewcommand{\arraystretch}{0.92}
\setlength{\tabcolsep}{2pt}
\begin{tabular*}{\columnwidth}{@{\extracolsep{\fill}}lcccc@{}}
\toprule
 & \multicolumn{2}{c}{External} & \multicolumn{2}{c}{Full regeneration} \\
\cmidrule(lr){2-3}\cmidrule(l){4-5}
Slot & Top-1 & Top-8 & Top-1 & Top-8 \\
\midrule
1 & 10.1 & 7.4 & 0.9 & 0.7 \\
2 & 6.0 & 7.6 & 0.7 & 1.2 \\
3 & 4.3 & 6.6 & 0.8 & 1.2 \\
4 & 2.9 & 5.4 & 0.7 & 0.9 \\
5 & 2.2 & 4.7 & 0.4 & 1.1 \\
6 & 1.3 & 4.1 & 0.6 & 0.8 \\
7 & 1.4 & 3.8 & 0.3 & 0.6 \\
\bottomrule
\end{tabular*}

\caption{Increments of KD over CE trained on ALLaVA, in points, in the top-1 agreement and the top-8 recall of the target's greedy token at each block slot, scored on the target's greedy text, averaged over two runs and 297 prompts.}
\label{tab:app-slots}
\end{table}

\section{Corpus Diagnostics}
\label{app:corpus}

Table~\ref{tab:app-corpus} reports how the target reads each version of the ALLaVA corpus. We run the target again over every retained example in bf16 with the training preprocessing, and each statistic follows the definition of the training gate, with each eligible anchor weighted by its probability of being sampled in training, $\min(128,n)/n$ for $n$ eligible anchors. Positions kept is the share of predicted slots that the gate keeps, and weight kept is the share of the position weight it keeps with $\gamma=7$ and with $\gamma=2$. The soft weight is the mean survival weight of PARD-2, and the corpus-token probability is the mean probability the target assigns to the corpus token. Intervals come from 10,000 resamples of examples.

\begin{table}[!htb]
\centering
\fontsize{7.5}{8.5}\selectfont
\renewcommand{\arraystretch}{0.92}
\setlength{\tabcolsep}{1pt}
\begin{tabular*}{\columnwidth}{@{\extracolsep{\fill}}lcccccc@{}}
\toprule
 & & & & & \multicolumn{2}{c}{Sampled} \\
\cmidrule(l){6-7}
Metric & External & Full & Trunc. & Window & $T{=}1$ & $T{=}0.7$ \\
\midrule
Agreement & 58.0 & 95.2 & 94.4 & 95.1 & 76.4 & 84.1 \\
Positions kept & 17.3 & 73.4 & 73.3 & 74.0 & 31.3 & 41.5 \\
Weight ($\gamma=7$) & 31.9 & 82.0 & 81.9 & 82.5 & 47.3 & 57.1 \\
Weight ($\gamma=2$) & 61.6 & 93.3 & 93.0 & 93.4 & 74.4 & 81.0 \\
Soft weight & 15.6 & 34.5 & 40.4 & 40.8 & 24.7 & 29.0 \\
\shortstack[l]{Corpus-token\\probability} & 52.4 & 79.2 & 82.3 & 82.7 & 68.4 & 74.1 \\
\bottomrule
\end{tabular*}

\caption{Corpus diagnostics over retained examples, in percent. Trunc. and Window denote truncation and the loss window. Agreement is the share of slots matching the target's greedy token. Every 95\% interval spans less than 0.3 points.}
\label{tab:app-corpus}
\end{table}

After full regeneration the remaining disagreement concentrates where the target is uncertain. Slots whose top-1 probability is below 0.5 hold 14.5\% of all slots but 74.1\% of the disagreements. Slots where it is at least 0.9 hold 49.7\% of all slots, and the corpus token differs from the greedy token at 0.06\% of them. Truncation inserts a closing token that the target never predicts at the cut, and closing tokens account for 14.1\% of the disagreements in the truncated version. Regenerated responses that reach the generation limit agree at 96.2\% against 95.2\% for the rest.

The sampled version regenerates every response once at temperature 1 with top-p 1, no top-k limit, at most 768 new tokens and one random seed shared across prompts, and it retains 26,195 examples. The second sampled version differs only in its temperature of 0.7 and also retains 26,195 examples.

Among the target-written versions, agreement does not track the KD gain in Table~\ref{tab:main}a: the matched-length versions agree within one point of full regeneration yet keep larger KD gains, and the sampled versions agree at only 76.4\% and 84.1\% yet keep KD gains close to the greedy one, although their gate gains lie between the external and greedy values. Nor does agreement rank the two corpora's external versions by KD gain: ShareGPT4V's external captions agree with the target more often than ALLaVA's external responses, yet KD gains more over CE on them.

\section{External Objective Settings}
\label{app:settings}

\begin{figure}[!htb]
\centering
\includegraphics[width=\columnwidth]{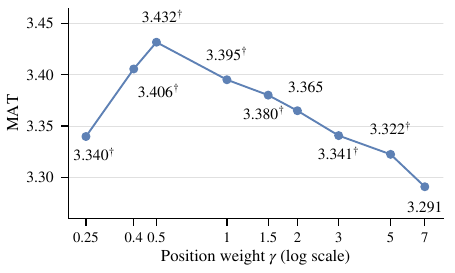}
\caption{Vision MAT of KD after three epochs on external ALLaVA responses as position weight $\gamma$ varies. $\dagger$ marks a single training run, and unmarked points average two runs.}
\label{fig:app-gamma}
\end{figure}

Table~\ref{tab:app-ladder} lists the objective settings trained on external responses, including one sampled evaluation at $T=1$, and Figure~\ref{fig:app-gamma} plots the position weight sweep for KD after three epochs. Entries marked $\dagger$ come from a single training run. The sampled evaluation lets the target sample at temperature 1 with random numbers fixed for each prompt and output position, shared by all drafters. Under sampled decoding the four settings keep their greedy-decoding order. KD with the soft survival weighting of PARD-2 in place of the gate, trained for three epochs with $\gamma=2$, reaches 3.542, the highest MAT among the external ALLaVA settings. Its gain over KD with $\gamma=2$ is 5.3\% in the main data order, and 4.6\% and 4.4\% when both are retrained once in each of the two further data orders.

\begin{table}[!htb]
\centering
\fontsize{7.5}{8.5}\selectfont
\renewcommand{\arraystretch}{0.92}
\setlength{\tabcolsep}{1.5pt}
\begin{tabular*}{\columnwidth}{@{\extracolsep{\fill}}lcccc@{}}
\toprule
 & \multicolumn{3}{c}{ALLaVA} & ShareGPT4V \\
\cmidrule(lr){2-4}\cmidrule(l){5-5}
Objective & 1 ep. & 3 ep. & 3 ep., $T=1$ & 3 ep. \\
\midrule
CE, $\gamma=7$ & 3.132 & 2.875 & 2.793 & 2.802 \\
KD, $\gamma=7$ & 3.400 & 3.291 & 3.180 & 3.307 \\
KD, $\gamma=2$ & 3.432 & 3.365 & 3.253 & 3.388$^\dagger$ \\
KD + gate, $\gamma=2$ & 3.530 & 3.526 & 3.362 & 3.602$^\dagger$ \\
\bottomrule
\end{tabular*}

\caption{Vision MAT of objective settings trained on external responses.}
\label{tab:app-ladder}
\end{table}

\section{Text Results}
\label{app:text}

Table~\ref{tab:app-text} lists task-level MAT of CE and KD for both drafters, Table~\ref{tab:app-text-gate} that of the gate and the label control, and Table~\ref{tab:cekd} gives overall MAT with intervals. Every fine-tuned DFlash drafter stays below its released initialization, trained on target-generated examples including math and code \citep{chen2026dflash}, and the losses concentrate on these tasks, which the chat corpus barely covers. On the two chat tasks, MT-Bench and Alpaca, the KD ratio of ratios is $-$8.2\% under greedy decoding and $-$8.4\% under sampling, a larger shrinkage than over all seven tasks, because under greedy decoding KD adds 10.1\% over CE on external responses there and 1.0\% on regenerated ones. These are the tasks where fine-tuning comes closest to the released drafter, and on Alpaca the regenerated CE and KD drafters are level with it while the ratio of ratios is $-$8.9\%.

\begin{table}[!htb]
\centering
\fontsize{7.5}{8.5}\selectfont
\renewcommand{\arraystretch}{0.92}
\setlength{\tabcolsep}{2pt}
\begin{tabular*}{\columnwidth}{@{\extracolsep{\fill}}lcccccc@{}}
\toprule
 & & & \multicolumn{2}{c}{External} & \multicolumn{2}{c}{Regenerated} \\
\cmidrule(lr){4-5}\cmidrule(l){6-7}
Task & $T$ & Init. & CE & KD & CE & KD \\
\midrule
\rowcolor{groupyellow}[0pt][0pt]\multicolumn{7}{@{}c@{}}{\textit{DFlash drafter}\strut} \\
\multirow{2}{*}{MT-Bench} & 0 & 3.25 & 2.43 & 2.65 & 3.08 & 3.11 \\
 & 1 & 3.12 & 2.35 & 2.55 & 2.96 & 2.95 \\
\multirow{2}{*}{Alpaca} & 0 & 2.71 & 2.12 & 2.35 & 2.71 & 2.74 \\
 & 1 & 2.62 & 2.07 & 2.32 & 2.56 & 2.62 \\
\multirow{2}{*}{GSM8K} & 0 & 6.06 & 3.82 & 4.09 & 5.16 & 5.20 \\
 & 1 & 5.60 & 3.67 & 3.91 & 4.83 & 4.89 \\
\multirow{2}{*}{AIME24} & 0 & 6.12 & 3.66 & 3.76 & 4.67 & 4.71 \\
 & 1 & 4.83 & 3.23 & 3.34 & 4.02 & 4.00 \\
\multirow{2}{*}{AIME25} & 0 & 6.59 & 3.88 & 4.01 & 5.12 & 5.17 \\
 & 1 & 5.08 & 3.53 & 3.54 & 4.29 & 4.34 \\
\multirow{2}{*}{HumanEval} & 0 & 6.32 & 4.09 & 4.47 & 5.10 & 5.18 \\
 & 1 & 5.63 & 3.79 & 4.15 & 4.65 & 4.77 \\
\multirow{2}{*}{LiveCodeBench} & 0 & 4.88 & 3.19 & 3.47 & 3.97 & 4.01 \\
 & 1 & 4.22 & 2.96 & 3.19 & 3.55 & 3.58 \\
\midrule
\multirow{2}{*}{Mean} & 0 & 4.89 & 3.23 & 3.46 & 4.14 & 4.18 \\
 & 1 & 4.29 & 3.02 & 3.23 & 3.75 & 3.79 \\
\rowcolor{groupyellow}[0pt][0pt]\multicolumn{7}{@{}c@{}}{\textit{EAGLE-3 drafter}\strut} \\
MT-Bench & 0 & 3.25 & 3.58 & 4.20 & 4.52 & 4.66 \\
Alpaca & 0 & 3.24 & 3.68 & 4.26 & 4.68 & 4.83 \\
GSM8K & 0 & 3.80 & 4.30 & 4.99 & 5.59 & 5.70 \\
AIME24 & 0 & 3.32 & 3.64 & 4.18 & 4.63 & 4.70 \\
AIME25 & 0 & 3.35 & 3.84 & 4.30 & 4.84 & 4.90 \\
HumanEval & 0 & 3.49 & 4.52 & 4.92 & 5.10 & 5.17 \\
LiveCodeBench & 0 & 3.26 & 3.74 & 4.17 & 4.29 & 4.41 \\
\midrule
Mean & 0 & 3.38 & 3.89 & 4.42 & 4.79 & 4.89 \\
\bottomrule
\end{tabular*}

\caption{Text-only MAT of CE and KD for each drafter, task and temperature. Init. is the released drafter, trained entries average two runs and Mean rows give geometric means over tasks. EAGLE-3 MAT comes from its own tree decoder.}
\label{tab:app-text}
\end{table}

For the DFlash drafter, the gate's ratio of ratios is negative in all seven tasks and in both training runs at either temperature. When the two runs of each cell are resampled along with the prompts, its interval is [$-$4.9, $-$3.6] under greedy decoding and [$-$6.0, $-$2.3] under sampling, and the interval of the gate's gain on regenerated responses includes zero at both temperatures. The label control of Section~\ref{sec:analysis} adds 7.4\% [6.9, 7.8] over CE on external and 0.3\% [0.1, 0.6] on regenerated responses under greedy decoding, a ratio of ratios of $-$6.5\% [$-$7.0, $-$6.1], or [$-$7.3, $-$5.8] when the two training runs of each cell are resampled along with the prompts. Under sampling it is $-$6.3\% [$-$7.6, $-$5.0], and at either temperature it is negative in all seven tasks and in both training runs. Under greedy decoding KD adds no measurable gain over greedy labels on external responses, 0.0\% [$-$0.3, 0.3], and 0.7\% [0.4, 1.0] on regenerated ones.

\begin{table}[!htb]
\centering
\fontsize{7.5}{8.5}\selectfont
\renewcommand{\arraystretch}{0.92}
\setlength{\tabcolsep}{1.5pt}
\begin{tabular*}{\columnwidth}{@{\extracolsep{\fill}}lccccc@{}}
\toprule
 & & \multicolumn{2}{c}{External} & \multicolumn{2}{c}{Regenerated} \\
\cmidrule(lr){3-4}\cmidrule(l){5-6}
Task & $T$ & KD + gate & \shortstack{Greedy\\labels} & KD + gate & \shortstack{Greedy\\labels} \\
\midrule
\multirow{2}{*}{MT-Bench} & 0 & 2.78 & 2.65 & 3.12 & 3.07 \\
 & 1 & 2.68 & 2.55 & 2.96 & 2.92 \\
\multirow{2}{*}{Alpaca} & 0 & 2.48 & 2.33 & 2.74 & 2.71 \\
 & 1 & 2.40 & 2.28 & 2.60 & 2.59 \\
\multirow{2}{*}{GSM8K} & 0 & 4.27 & 4.10 & 5.22 & 5.17 \\
 & 1 & 4.09 & 3.93 & 4.89 & 4.83 \\
\multirow{2}{*}{AIME24} & 0 & 3.82 & 3.79 & 4.72 & 4.71 \\
 & 1 & 3.41 & 3.36 & 3.99 & 4.01 \\
\multirow{2}{*}{AIME25} & 0 & 4.23 & 4.09 & 5.18 & 5.18 \\
 & 1 & 3.68 & 3.61 & 4.34 & 4.27 \\
\multirow{2}{*}{HumanEval} & 0 & 4.75 & 4.42 & 5.17 & 5.13 \\
 & 1 & 4.43 & 4.15 & 4.75 & 4.72 \\
\multirow{2}{*}{LiveCodeBench} & 0 & 3.64 & 3.46 & 4.01 & 3.96 \\
 & 1 & 3.32 & 3.17 & 3.58 & 3.56 \\
\midrule
\multirow{2}{*}{Mean} & 0 & 3.62 & 3.46 & 4.19 & 4.16 \\
 & 1 & 3.36 & 3.23 & 3.78 & 3.76 \\
\bottomrule
\end{tabular*}

\caption{Text-only MAT of the DFlash drafter with the survival gate and with greedy labels for each task and temperature, averaged over two training runs. Greedy labels trains CE toward the target's greedy token. Mean rows give geometric means over tasks.}
\label{tab:app-text-gate}
\end{table}

For the EAGLE-3 drafter, KD adds 13.7\% [13.2, 14.3] over CE on external and 2.2\% [1.8, 2.5] on regenerated responses, and the ratio of ratios is $-$10.2\% [$-$10.7, $-$9.6], or [$-$11.0, $-$9.3] when the two runs of each cell are resampled along with the prompts. In the two data orders it is $-$10.7\% and $-$9.7\%, and across tasks it ranges from $-$12.1\% on GSM8K to $-$6.9\% on HumanEval, with every task's interval below zero, also with the runs resampled. Every fine-tuned EAGLE-3 checkpoint exceeds the released drafter in every task, and its MAT over all tasks is 14.5\% to 44.7\% above the released 3.383.

\end{document}